\pdfoutput=1
\documentclass{article}

\PassOptionsToPackage{numbers}{natbib}

\usepackage[preprint]{neurips_2023}

\usepackage[utf8]{inputenc} 
\usepackage[T1]{fontenc}    
\usepackage{hyperref}       
\usepackage{url}            
\usepackage{booktabs}       
\usepackage{amsfonts}       
\usepackage{nicefrac}       
\usepackage{microtype}      
\usepackage[ruled]{algorithm2e}

\usepackage{graphicx}
\usepackage{subfigure}
\usepackage{booktabs} 
\usepackage{caption}

\usepackage{framed}
\usepackage{amssymb}
\usepackage{amsfonts}
\usepackage{mathrsfs}
\usepackage{mathtools}
\usepackage{array}
\usepackage{amsthm}
\usepackage{verbatim} 
\usepackage{enumerate}
\usepackage{bbm}
\usepackage{commath}
\usepackage{wrapfig}
\usepackage{amsbsy}
\usepackage{float}
\usepackage{amsmath}
\usepackage{tabularx}
\usepackage{listings}
\usepackage{xcolor}
\usepackage[shortlabels]{enumitem}
\usepackage{tcolorbox}
\usepackage{multirow}
\usepackage{CJKutf8}
\usepackage{url}

\definecolor{codegreen}{rgb}{0,0.3,0.6}
\definecolor{codegray}{rgb}{0.5,0.5,0.5}
\definecolor{codepurple}{rgb}{0.58,0,0.82}
\definecolor{backcolour}{rgb}{0.95,0.95,0.92}

\definecolor{darkblue}{rgb}{0.0,0.0,0.66}   
\hypersetup{colorlinks=true,linkcolor=red,citecolor=darkblue}

\lstdefinestyle{mystyle}{
    basicstyle=\tiny,
    commentstyle=\color{codegreen},
    keywordstyle=\color{magenta},
    numberstyle=\tiny\color{codegray},
    stringstyle=\color{codepurple},
    basicstyle=\fontsize{8.5}{9}\selectfont\ttfamily,
    breakatwhitespace=false,         
    breaklines=true,                 
    captionpos=b,                    
    keepspaces=true,                 
    numbers=left,                    
    numbersep=5pt,                  
    showspaces=false,                
    showstringspaces=false,
    frame = single
}

\newcommand{\model}{MathGLM\xspace}

\newcommand{\vpara}[1]{\vspace{0.07in}\noindent\textbf{#1 }}

\title{GPT Can Solve Mathematical Problems Without a Calculator}

\author{%
    Zhen Yang$^{\dag*}$, Ming Ding$^{\S\dag*}$, Qingsong Lv$^\S$, Zhihuan Jiang$^\dag$, Zehai He$^\dag$, Yuyi Guo$^\dag$, \\ \bf{Jinfeng Bai$^\diamond$, Jie Tang$^{\dag\ddag}$} \\  \\
    $^\dag$Tsinghua University \hspace{0.3cm}  $^\diamond$TAL AI Lab \hspace{0.3cm}  $^\S$Zhipu.AI 
}

\begin{document}

\maketitle

\renewcommand{\thefootnote}{\fnsymbol{footnote}}
    \footnotetext[1]{ZY and MD contributed equally (\texttt{\{yangz21,dm18\}@mails.tsinghua.edu.cn}).}
    \footnotetext[3]{Corresponding authors: Jie Tang (\texttt{jietang@tsinghua.edu.cn})}

\newtheorem{thm}{Theorem}
\newtheorem{definition}[thm]{Definition}
\newtheorem{lemma}[thm]{Lemma}
\newtheorem{theorem}[thm]{Theorem}
\newtheorem{corollary}[thm]{Corollary}
\newtheorem{remark}[thm]{Remark}
\definecolor{comment}{RGB}{70, 150, 60}

\begin{abstract}
Previous studies have typically assumed that large language models are unable to accurately perform arithmetic operations, particularly multiplication of >8 digits, and operations involving decimals and fractions, without the use of calculator tools. This paper aims to challenge this misconception. With sufficient training data, a 2 billion-parameter language model can accurately perform multi-digit arithmetic operations with almost 100\% accuracy without data leakage, significantly surpassing GPT-4 (whose multi-digit multiplication accuracy is only 4.3\%). We also demonstrate that our MathGLM, fine-tuned from GLM-10B on a dataset with additional multi-step arithmetic operations and math problems described in text, achieves similar performance to GPT-4 on a 5,000-samples Chinese math problem test set. Our code and data are public at \url{https://github.com/THUDM/MathGLM}.

\end{abstract}

\section{Introduction}

\begin{figure}[htbp]
    \centering
    \includegraphics[width=0.7\textwidth]{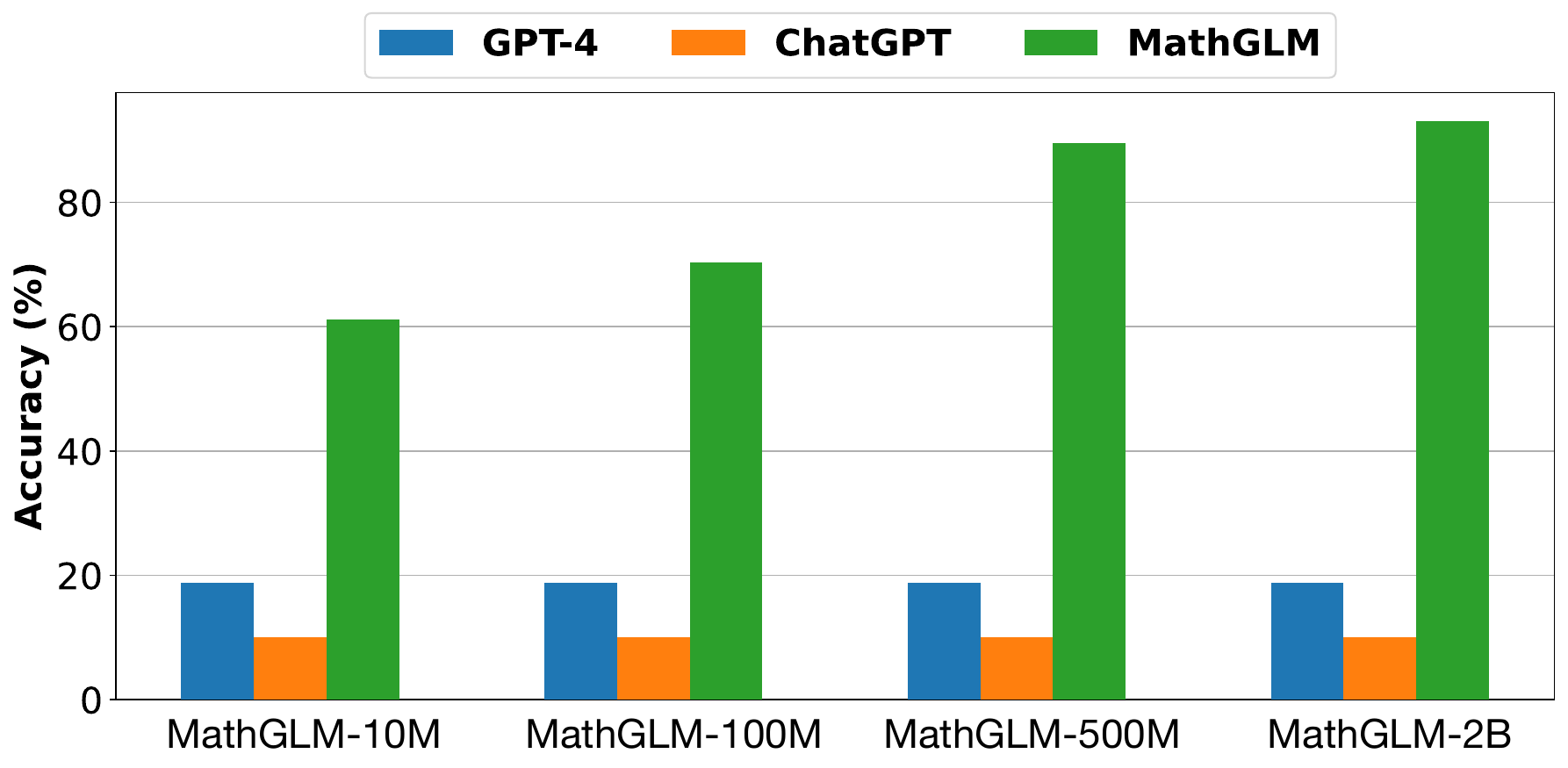}
    \caption{Accuracy scores across various LLMs like GPT-4 and ChatGPT, as well as a series of \model models on the generated test dataset for the arithmetic tasks. Among the different model scales, \model consistently achieves superior performance. 
    }
    \label{fig:arithmetic_performance}
\end{figure}

Large language models (LLMs) have demonstrated remarkable ability in handling a variety of downstream tasks in the NLP domain~\cite{brown2020language,chowdhery2022palm,zeng2022glm,thoppilan2022lamda,zhang2022opt,scao2022bloom}. Pioneering models, such as GPT-4~\cite{2303.08774} and ChatGPT~\cite{chatgpt}, have been trained on massive amounts of text data, enabling them to generate coherent and contextually relevant responses. Their ability to understand and generate text makes them highly versatile for various NLP tasks. Moreover, LLMs have been leveraged for other assignments, involving areas of mathematics~\cite{cobbe2021training,lewkowycz2022solving} and science~\cite{taylor2022galactica}. Nevertheless, despite the impressive capabilities across diverse NLP tasks, GPT-4 might not exhibit the same level of proficiency in mathematical reasoning, including arithmetic tasks and Chinese math word problems.

\begin{figure}[htbp]
    \centering
    \includegraphics[width=0.98\textwidth]{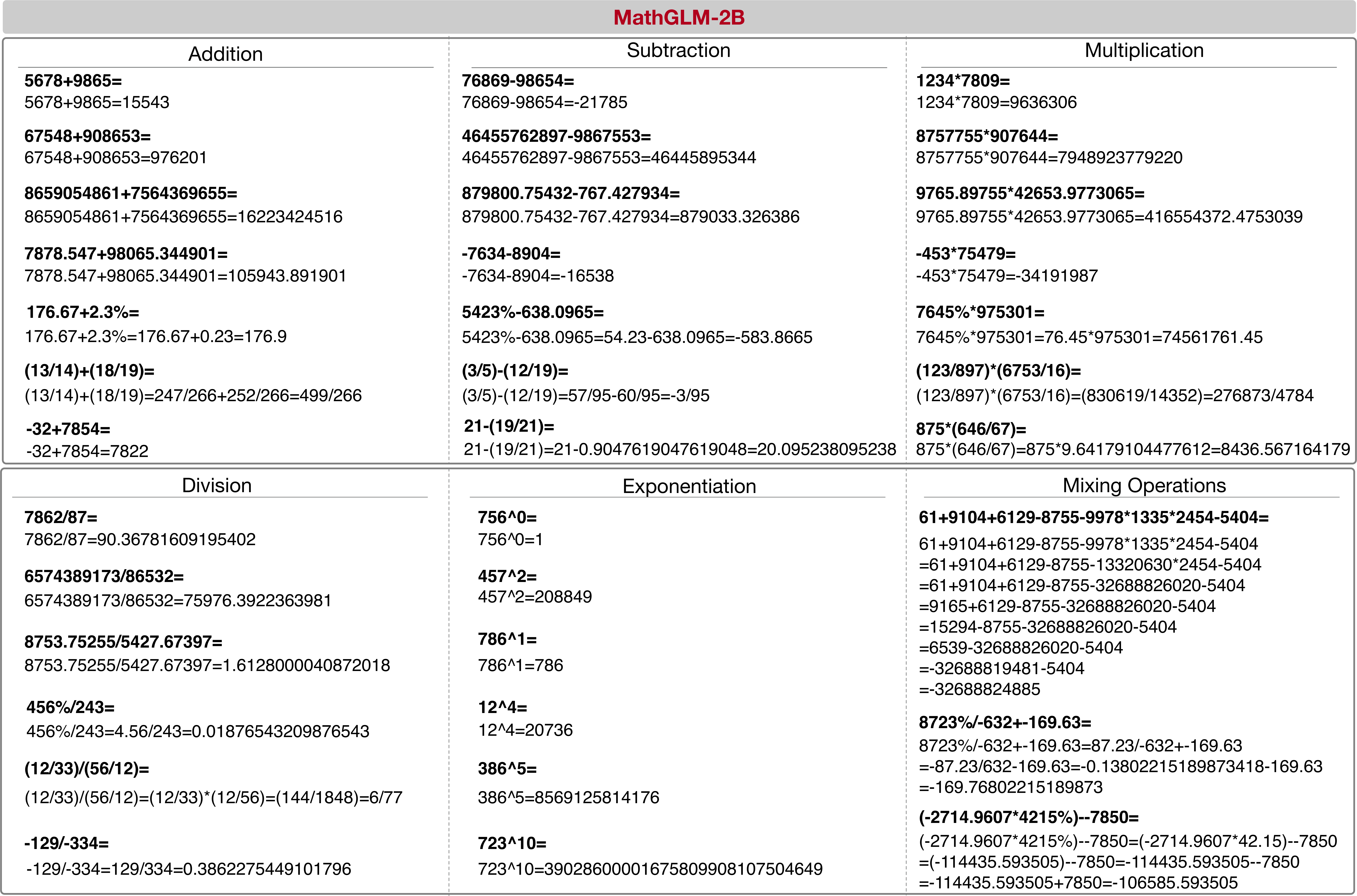}
    \caption{Examples of \model's response on a variety of arithmetic tasks.
    }
    \vspace{-0.6cm}
    \label{fig:arithmetic_exampls}
\end{figure}

In the context of arithmetic tasks, a prevailing assumption is that LLMs struggle with accurately executing complex arithmetic operations,  especially pronounced in cases involving multiplication of numbers exceeding 8 digits, and operations entailing decimals and fractions. To eliminate these misconceptions, we embark on an investigation to assess the arithmetic ability of LLMs. Specifically, we focus on the capability of LLMs in performing complex arithmetic operations. As a result, we propose \model, a powerful model meticulously crafted to impeccably execute an extensive spectrum of complex arithmetic operations, achieving the best performance compared to leading LLMs such as GPT-4 (See Figure~\ref{fig:arithmetic_performance}). These operations contain singular actions like addition, subtraction, multiplication, division, and exponentiation, as well as the mixing of these operations employing brackets. 
When these operations are performed individually, without being combined with any other operation, we refer to them as ``1-atomic operation''. Importantly, \model has the capability to adeptly tackle arithmetic operations that involve a variety of numerical forms, including integers, decimals, fractions, percentages, and even negative numbers. 
Figure~\ref{fig:arithmetic_exampls} demonstrates examples generated by \model with 2B model parameters on addition, subtraction, multiplication, division, exponentiation, and mixing operations tasks.

To attain the remarkable performance exhibited by \model in arithmetic tasks, we utilize a step-by-step strategy to construct an arithmetic dataset that serves as the foundation for \model's pre-training. This dataset is designed to encompass a wide spectrum of arithmetic operations, spanning from straightforward 1-atomic operation to more complex 9-atomic operations. By adopting this step-by-step strategy, \model learns to handle both simple and intricate arithmetic expressions, which empowers it to accurately perform calculations even for operations involving multiplication of numbers greater than 8 digits, and those with decimals and fractions. Moreover, we incorporate the concept of curriculum learning to further augment the capabilities of \model. By gradually increasing the complexity of the arithmetic expressions, \model progressively enhances its capacity to tackle operations involving numbers spanning up to 12 digits. This stands in contrast to the common assumption that large language models struggle with such complex arithmetic tasks. The results demonstrate that \model's arithmetic performance surpasses even the most robust LLMs like GPT-4. Specifically, \model achieves an impressive accuracy of 93.03\% on the test dataset containing complex mixed operations. In contrast, GPT-4 only manages a meager 18.84\% accuracy on the same dataset.

For math word problems, the Ape210K dataset~\cite{zhao2020ape210k} serves as a comprehensive source of mathematical challenges, drawing from diverse math word problems across the Internet. This dataset serves as a valuable resource for training \model, offering a broad spectrum of problem types for learning. However, a notable characteristic of the original dataset lies in its directly calculated answers. This straightforward answer presentation might lead to a potential drawback, that is \model can potentially miss the underlying calculation rules and patterns embedded within the calculation processes.

To overcome this potential limitation and bolster \model's proficiency in solving math word problems,
we leverage the step-by-step strategy to reconstruct the Ape210K dataset. By decomposing the complex arithmetic calculation process into a sequence of sequential steps, \model is empowered to accurately generate answer for math word problems and significantly enhance the answer accuracy in comparison to the original one. For instance, \model achieves an impressive absolute gain of 42.29\% in answer accuracy as compared to fine-tuning on the original dataset. By fine-tuning from the GLM-10B, \model's performance closely aligns with that of GPT-4 when evaluated on a math word problems dataset comprising 5,000 test cases. This step-by-step strategy provides \model with a profound understanding of the complex calculation process inherent in math word problems, enabling \model to grasp the underlying calculation rules and obtain more accurate answers.

Overall, \model excels in both arithmetic tasks and math word problems by leveraging the step-by-step strategy. Our comprehensive experiments and detailed analysis demonstrate the effectiveness of \model's mathematical reasoning compared to GPT-4. These results significantly challenge the common misconception that LLMs struggle with complex arithmetic tasks, thus unveiling their remarkable potential to excel in the realm of mathematical reasoning tasks. We organize this paper as follows. In Section~\ref{sec:related_work}, we elaborate on preliminaries, including large language models, arithmetic calculation, and mathematical reasoning. Section~\ref{sec:method} introduces the methodologies employed in \model, covering arithmetic training dataset, models, and training procedure for arithmetic tasks (Section~\ref{subsec:arithmetic}), and training dataset, backbone models, and training strategy for math word problems (Section~\ref{subsec:mwp}). We also perform comprehensive experiments and an analysis of the \model's capabilities (Section~\ref{sec:sec_exp}). Section~\ref{subsec:exp_arithmetic} reports the detailed experimental results on arithmetic tasks, and Section~\ref{subsec:exp_mwp} presents the results related to math word problems. Finally, we summarize our work in Section~\ref{sec:conclusion}.

\section{Related Work}\label{sec:related_work}
\subsection{Large Language Models}
Large Language Models (LLMs) have demonstrated robust capabilities in the realm of Natural Language Processing (NLP) tasks, significantly shifting the research paradigm within the field. These models, such as GPT-3~\cite{brown2020language}, Gopher~\cite{rae2021scaling}, Megatron-Turing NLG~\cite{smith2022using}, Chinchilla~\cite{hoffmann2022training}, PaLM~\cite{chowdhery2022palm}, OPT~\cite{zhang2022opt}, BLOOM~\cite{scao2022bloom}, GLM-130B~\cite{zeng2022glm}, and LLaMA~\cite{touvron2023llama}, are trained on a large corpus of diverse and unlabeled data, demonstrating a generic ability to perform well on a variety of tasks. Through pretraining on extensive corpus, these models obtain powerful language understanding and generation capabilities, enabling their exceptional performance on a wide array of benchmarks, such as MMLU~\cite{hendrycks2020measuring}, mathematical reasoning, and code generation. Moreover, they display an astonishing aptitude for in-context learning, rapidly adapting to novel tasks with minimal examples through few-shot learning.

Nonetheless, despite the remarkable strides made by the most powerful LLMs, ChatGPT~\cite{ouyang2022training} and GPT-4~\cite{2303.08774}, in language understanding and generation, it is crucial to recognize that these cutting-edge models still encounter challenges in tackling mathematical problems. This work is dedicated to addressing and enhancing the performance of LLMs in the domain of solving mathematical problems, encompassing both arithmetic tasks and math word problems.

\subsection{Arithmetic Calculation}
The emergence of pre-trained Large Language Models (LLMs)~\cite{brown2020language,chowdhery2022palm,2303.08774} has sparked considerable interest in investigating their potential for handling arithmetic tasks. ~\citet{nogueira2021investigating} and ~\citet{wang2021exploring} evaluate the arithmetic capabilities of LLMs on  elementary arithmetic operations like addition and subtraction. ~\citet{muffo2023evaluating} undertake an evaluation that specifically centers on assessing the proficiency of language models in the domain of 2-digit multiplication. BIG-bench~\cite{srivastava2022beyond} introduces a comprehensive collection of arithmetic datasets, which encompass a spectrum of arithmetic tasks that span numbers within a range of up to 5 digits. ~\citet{yuan2023well} design an complex arithmetic dataset MATH 401 with various arithmetic operations to evaluate the capabilities of models like GPT-4, ChatGPT, InstructGPT~\cite{ouyang2022training}, Galactica~\cite{taylor2022galactica}, and LLaMA~\cite{touvron2023llama}. 

To support arithmetic operations involving large numbers, ~\citet{nye2021show} employ scratchpad-based fine-tuning that enables LLMs to achieve remarkable outcomes in the context of 8-digit addition.  ~\citet{zhou2022teaching} adopt the specialize prompt engineering techniques to successfully extend the scope of addition but encountered limitations with multiplication beyond 7 digits. Goat~\cite{liu2023goat} utilizes supervised instruction fine-tuning to handle elementary arithmetic operations with large integers, including addition, subtraction, multiplication, and division. ~\citet{jelassi2023length} investigate length generalization in basic arithmetic tasks via approaches like relative position embeddings and train set priming. Distinguishing itself from these efforts focused on elementary arithmetic, our \model pushes the envelope by not only exceeding the realm of basic arithmetic with two numbers but also tackling intricate mixing arithmetic operations involving multiple numbers and diverse data formats. 

Furthermore, several works explore the integration of external tools for arithmetic tasks. For instance, Toolformer~\cite{schick2023toolformer} adopts an external calculator to accomplish arithmetic calculations, while PoT~\cite{chen2022program} and PAL~\cite{gao2023pal} obtain the final answer with the help of programs. Different from leveraging external tools, we focus on explore how to enhance the inherent arithmetic ability of LLMs without relying on external tools.

\subsection{Mathematical Reasoning}
LLMs have indeed demonstrated considerable promise in addressing math word problems. 
~\citet{cobbe2021training} utilize training verifiers to rerank the outputs of LLMs, resulting in remarkable performance on the created GSM8K dataset. ~\citet{lewkowycz2022solving} introduce Minerva, a large language model fine-tuned based on PaLM models~\cite{chowdhery2022palm}, leveraging a substantial dataset containing scientific and mathematical data. Minerva attains state-of-the-art performance on MATH~\cite{hendrycks2021measuring} and GSM8K. By leveraging \textit{COT (chain of thought)}~\cite{wei2022chain,kojima2022large,zhou2022least} to decompose the math problems into multiple steps, LLMs notably improve their performance in tackling math word problems. ~\citet{wang2022self} propose the \textit{self-consistency} strategy as a replacement for the decoding strategy used in COT, which brings about better performance than the traditional COT prompting. ~\citet{uesato2022solving} employ process and outcome supervision to enhance the performance of LLMs in solving grade school math problems. ~\citet{2305.20050} propose to verify each intermediate reasoning step and find process supervision can significantly improve mathematical reasoning performance. While these studies show the substantial advancements made by LLMs in mathematical reasoning, it is clear that LLMs still make mistakes when confronted with arithmetic operations in math word problems. Different from the aforementioned works that primarily concentrate on improving the reasoning process, our goal is to simultaneously advance both mathematical reasoning and arithmetical calculation capabilities of LLMs, addressing both aspects at the same time.

\section{Method}\label{sec:method}

To investigate the efficacy of LLMs in mathematical reasoning, we propose the \model model that designed with the specific goal of enhancing the performance of LLMs in mathematical reasoning. \textbf{Firstly}, \model focuses on enhancing its proficiency in accurately executing a comprehensive range of arithmetic tasks. It accomplishes this by integrating a step-by-step strategy into its architecture. Instead of straightforwardly calculating the answers to complex arithmetic expressions, \model employs this strategy to meticulously generate answers step by step. \textbf{Secondly}, \model leverages the step-by-step strategy to fine-tune a series of GLM models on specific Chinese mathematical problems. By leveraging this strategy, \model enhances its ability to handle complex mathematical problem-solving tasks.

\subsection{Learning on Arithmetic Tasks}\label{subsec:arithmetic}

Arithmetic tasks can be broadly divided into basic arithmetic operations and complex mixing operations. Basic arithmetic operations encompass fundamental mathematical tasks that revolve around conducting simple calculations involving two numbers. On the other hand, arithmetic tasks also encompass the domain of complex mixing operations, which necessitate the skill to manage a combination of diverse arithmetic operations and numerical formats. A comprehensive category of the learning tasks encompassed by \model is summarized in Table~\ref{tab:learning_task}.

\begin{small}
\begin{table*}[hbpt]
    \centering
    \renewcommand{\arraystretch}{1.15}
    \setlength{\tabcolsep}{0.8mm}{
    \begin{tabular}{l|c|c|c|c|c}  
    \toprule
    \multirow{1}{*}{Task} & \multicolumn{1}{c|}{Integer}  & \multicolumn{1}{c|}{Decimal} & \multicolumn{1}{c|}{Fraction} & \multicolumn{1}{c|}{Percentage} & \multicolumn{1}{c}{Negative Numbers}  \\
    \midrule 
    Addition & nD+nD  & nD.mD+nD.mD & (nD/mD)+(nD/mD) & nD\%+nD\% & -nD+-nD  \\
    Subtraction & nD-nD  & nD.mD-nD.mD & (nD/mD)-(nD/mD) & nD\%-nD\% & -nD--nD  \\
    Multiplication & nD*nD & nD.mD*nD.mD & (nD/mD)*(nD/mD) & nD\%*nD\% & -nD*-nD \\
    Division & nD/nD & nD.mD/nD.mD & (nD/mD)/(nD/mD) & nD\%/nD\% & -nD/-nD   \\
    Exponentiation &  nD$\textasciicircum$nD & - & - & - & -nD$\textasciicircum$-nD \\
    \midrule
    \midrule
    Mixed Computing & \multicolumn{5}{c}{[(nD$\pm$nD.mD)*nD\%]/-nD}  \\ 
    \bottomrule
    \end{tabular}} 
    \caption{Summary and symbolic expression of arithmetic tasks. In symbolic expression, we represent a decimal with n-digit integer part and m-digit decimal part as nD.mD. For mixed computing, we only show a simple mixed symbolic expression. }
    \vspace{-0.2cm}
    \label{tab:learning_task}
\end{table*}
\end{small}

To augment the arithmetic ability of \model, we adopt a decoder-only architecture based on Transformer~\cite{vaswani2017attention} and train it from scratch on our generated arithmetic dataset using an autoregressive objective. 

\begin{figure}[htbp]
    \centering
    \includegraphics[width=\textwidth]{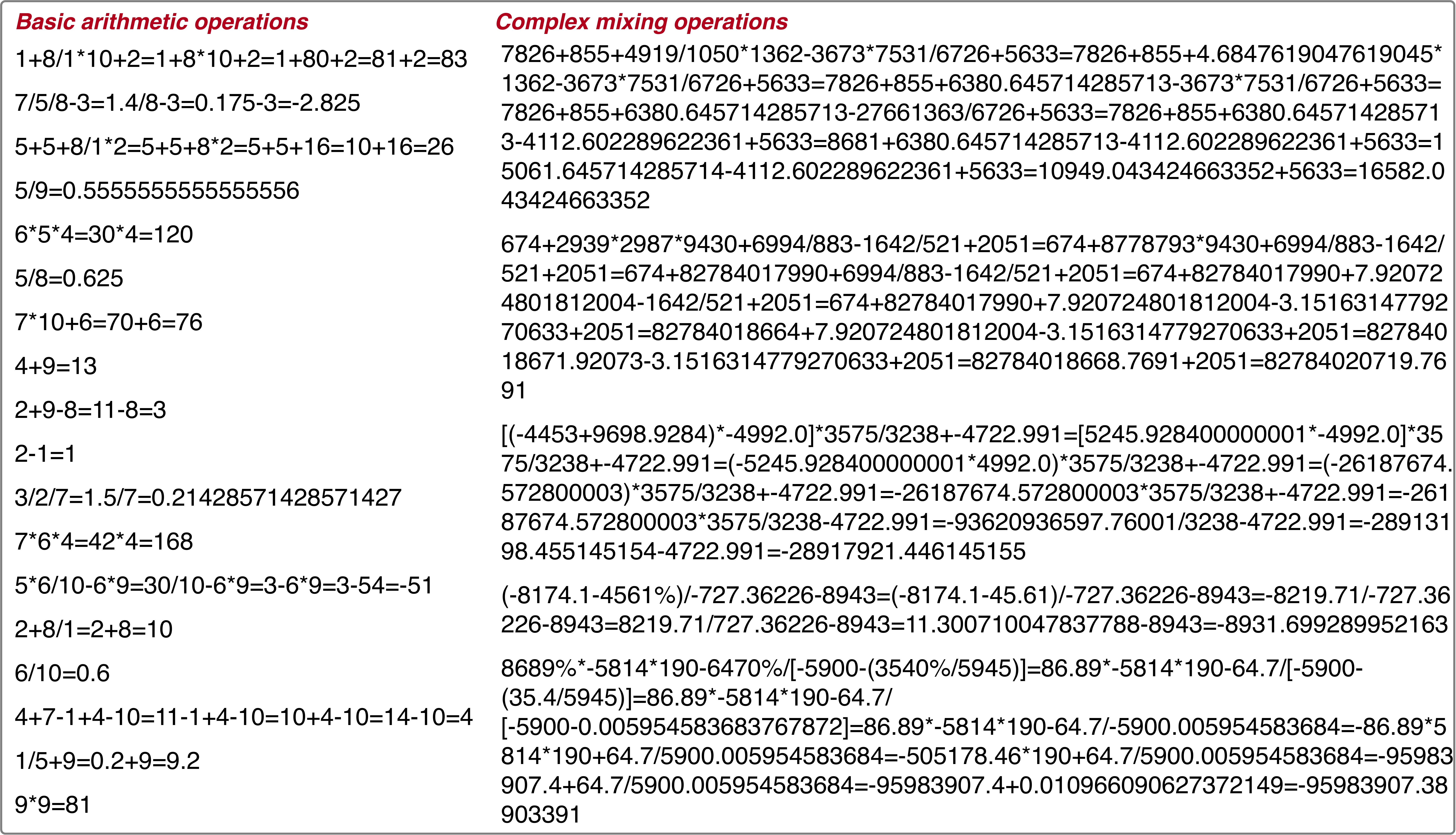}
    \caption{Some examples of the arithmetic training dataset of \model.
    }
    \vspace{-0.2cm}
    \label{fig:arithmetic_dataset}
\end{figure}

\vpara{Arithmetic Training Dataset.} The arithmetic dataset employed for training is meticulously designed to encompass a comprehensive range of arithmetic tasks. This dataset is thoughtfully designed to incorporate a variety of operations, including addition, subtraction, multiplication, division, and exponentiation. Additionally, it encompasses diverse numerical formats such as integers, decimals, percents, fractions, and negative numbers. This comprehensive dataset is created in various sizes, ranging from 1 million to 50 million records. Within each of these datasets, individual arithmetic expressions consist of 2 to 10 operation steps, encompassing a spectrum of mathematical operations like addition (+), subtraction (-), multiplication ($\times$), division (/), and exponentiation ($\textasciicircum$). To aligh with human calculation habits, a step-by-step strategy is employed in the construction of the arithmetic datasets. Instead of directly computing the final answer to each complex arithmetic expression, the strategy breaks down the complex expression into a sequence of simpler steps, progressively generating answers step by step. This strategy mirrors the process human typically follow when solving complex arithmetic tasks. By training on such dataset, \model achieves outstanding arithmetic performance since it learns the underlying calculation rules from the detailed calculation process. Figure~\ref{fig:arithmetic_dataset} provides some training examples drawn from the arithmetic dataset, illustrating the diversity of arithmetic tasks and the step-by-step strategy incorporated in the dataset.

\vpara{Models and Training Procedure.} 
Table~\ref{tab:model_size} reports an overview of all the models with different model parameters. Our training efforts encompass 4 distinct types of models, each characterized by different parameter sizes. The largest model is endowed with 2B parameters, making it the most powerful in terms of capacity. Following that, we train the second model with 500M parameters, the third model with 100M parameters and the smallest model with 10M parameters. Notably, despite the discrepancies in parameter sizes, all models are trained using the same dataset scale consisting of 50 million training records. The technical details of \model about tokenization is presented in Appendix~\ref{appendix:arithmetical_hyperparameters}.

\begin{small}
    \begin{table}[h]
    \centering
    \renewcommand{\arraystretch}{1.15}
    \setlength{\tabcolsep}{1.5mm}{
    \begin{tabular}{c|ccccc}  
    \toprule
     Model     & Dimension   & Heads & Layers & Parameters & Training Steps   \\
    \midrule
    \model-10M & 256 & 32 & 15 & 10M  & 120,000 \\
    \model-100M  & 512 & 32 & 35 & 100M & 155,000 \\
    \model-500M  & 1024& 32 & 40 & 500M & 135,000  \\
    \model-2B    & 2048& 32 & 40 & 2B   & 155,000  \\
    \bottomrule
    \end{tabular}} 
    \vspace{1mm}
    \caption{Model sizes and architectures of \model.}
    \vspace{-0.5cm}
    \label{tab:model_size}
    \end{table}
\end{small} 

For training procedure, we employ the fundamental principle of curriculum learning to effectively train the \model. The training procedure of \model is initiated using an arithmetic dataset containing numbers within a range of 5 digits. Following this initial phase, where \model attains stable training convergence and demonstrates satisfactory performance on the test dataset, we introduce curriculum learning to enhance its capabilities. Specifically, we augment the training data with a new dataset comprising 50,000 records, which encompass numbers spanning from 5 to 12 digits. By incorporating these more challenging examples, \model is encouraged to decipher the rules associated with arithmetic operations involving large numbers. Such training strategy allows \model initially tackles simpler examples, progressively advancing towards more complex challenges. More importantly, such approach empowers \model to improve its ability by learning from relatively smaller examples, emphasizing the efficiency of \model to handle increasingly intricate tasks or data patterns.

\subsection{Learning on Math Word Problems}\label{subsec:mwp}

Alongside our focus on arithmetic tasks, we train (fine-tune) a series of Transformer-based language models, named General Language Model (GLM)~\cite{du2021glm,zeng2022glm} and their chat versions to solve math word problems. Our training leverages the publicly available Chinese Ape210K dataset, which serves as a valuable resource for training language models on math word problem-solving tasks. This dataset consists of a vast collection of 210,000 Chinese math problems at the primary school level, with each problem's answer calculated directly.

\begin{figure}[htbp]
    \centering
    \includegraphics[width=\textwidth]{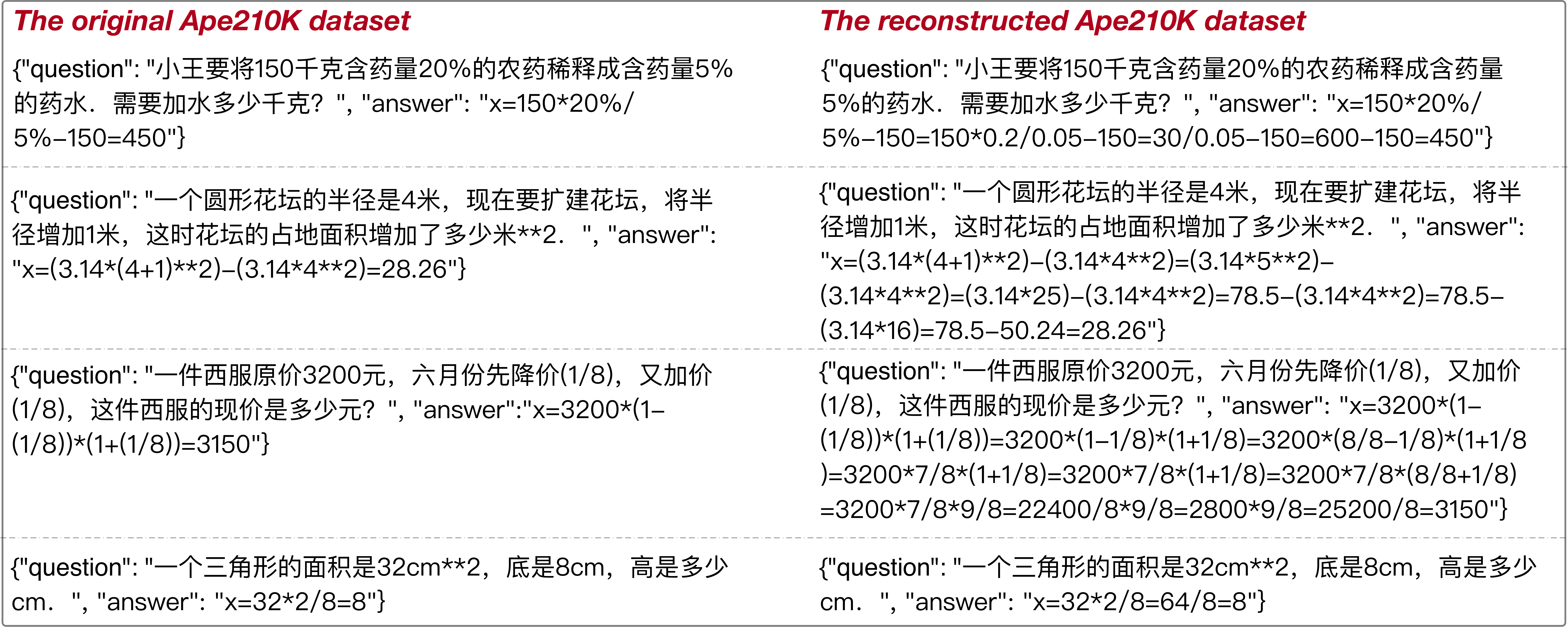}
    \caption{Comparison between the original Ape210k dataset and the reconstructed version. A step-by-step strategy is employed to reconstruct the solutions for each mathematical problem.
    }
    \vspace{-0.5cm}
    \label{fig:data_mwp}
\end{figure}

\vpara{Training Dataset.} To enhance the performance of \model on math word problems, we utilize a step-by-step strategy to reconstruct the Ape210K dataset, transforming it into a version where the answer of each math problem is calculated step by step. Figure~\ref{fig:data_mwp} demonstrate the contrast between the original Ape210K dataset and our reconstructed version. The newly reconstructed dataset encourages \model to acquire an in-depth understanding of the underlying calculation rules inherent in solving math word problems. Through this step-wise process, \model becomes adept at deriving a final, accurate answer for each problem, emphasizing its ability to harness the complexities of mathematical reasoning.

\vpara{Backbone Models.} We adopt different variations of the GLM as the backbone to train the \model, including GLM-large with 335M parameters, GLM-6B, GLM2-6B, and GLM-10B. Besides, we train the \model using the ChatGLM-6B and ChatGLM2-6B backbones. These backbone models bestow the \model with a basic language understanding skills, enabling it to effectively comprehend linguistic information contained within math word problems. The details of backbone models are presented in Appendix~\ref{appendix:backbone_models}.

\vpara{Training Strategy.} To achieve better performance, we employ two training strategies for \model. The first is to fine-tune the GLM backbone models on a solitary mathematical dataset. This process allows the \model to specialize in understanding and solving math word problems by learning from the mathematical dataset's unique characteristics. However, such strategy damages the generic ability of the \model. To circumvent this limitation, a second strategy is to continue training the GLM backbone models on a hybrid dataset that combines both mathmatics and text content. This helps to balance the specialization in math word problems with the preservation of \model's generic ability.

 \section{Experiments}
\label{sec:sec_exp}

The overarching objective of \model revolves around demonstrating the prowess of language models in the domain of mathematical reasoning. To validate this, we design two distinct types of experiments, encompassing arithmetic tasks and math word problems. These two categories of tasks comprehensively cover both basic computational abilities and higher-order problem-solving skills, providing a robust assessment of the model's proficiency in mathematical reasoning.

\subsection{Learning on Arithmetic}\label{subsec:exp_arithmetic}
\subsubsection{Dataset}
Within the domain of arithmetic, we create a diverse collection of datasets specifically tailored for arithmetic tasks. This suite of training datasets encompasses an expansive spectrum of sizes, including 1 million, 5 million, 10 million, 25 million and 50 million records. Our evaluation dataset, which comprises 9,592 test cases, is generated from the same distribution as the training dataset, yet remains distinct and is excluded from the training process. This carefully generated suite of datasets serves as a comprehensive benchmark to evaluate and quantify \model's computational prowess across a wide variety of arithmetic tasks. For a more in-depth exploration of the specifics of the generated datasets, the details can be found in Appendix~\ref{appendix: dataset}.

\subsubsection{Evaluation Metric}
To measure the ability of \model on arithmetic tasks, we adopt the following metrics to evaluate the outputs.

\vpara{Accuracy} is typically measured by comparing the output of the \model and the ground truth answer. In our experiments, we adhere to standard rounding rules, constraining the generated answers to precisely two decimal places. When the correctly rounded answer aligns with the answer generated by the \model, we classify this outcome as a correct answer.

\vpara{Relative Error} is another important metric used to evaluate the effectiveness of \model, which quantifies the difference between the output generated by \model and the correct answer. The relative error (RE) is quantified using the following formula:
\begin{equation}
    RE = |\frac{\hat{y} - y}{y}|
\end{equation}
where $\hat{y}$ and $y$ denote the generated answer and the correct answer respectively. For our evaluation purposes, we utilize a relative error threshold of 1\%. This threshold serves as a criterion for determining the acceptability of the answers generated by the \model, where any relative error falling within this threshold range is considered an accurate outcome.

\subsubsection{Results and Analysis}

\vpara{Overall Results.} For arithmetic tasks, we pre-train a Transformer-based model named \model with 500M model parameters for both pretraining and inference. To accurately gauge the effectiveness of \model, we contrast its performance with those of leading large language models (LLMs) such as GPT-4 and ChatGPT. The results, as presented in Table~\ref{tab:arithmetic_results}, consistently show that \model outperforms all other models, indicating its superior performance in tackling arithmetic tasks. Even when we consider a more small model variant, namely \model-10M with a mere 10 million parameters, the results reveal a surprising phenomenon. Despite its compact parameter size, \model-10M outperforms GPT-4 and ChatGPT across an array of comprehensive arithmetic tasks. This astonishing results show the effectiveness of \model's approach, which involves decomposing complex arithmetic expressions into individual steps, granting it the capacity to discern and comprehend the subtleties within arithmetic tasks. It effectively learns the underlying rules and principles of arithmetic operations, enabling it to generate accurate and precise solutions. Furthermore, when comparing \model across different parameter scales, we observe that the \model's arithmetic performance is directly correlated with the augmentation of its parameter count. This finding suggest that as models increase in size, their performance exhibits a corresponding enhancement. To sum up, the evaluation results on complex arithmetic tasks underscore the exceptional performance of \model. By breaking down arithmetic tasks, these models surpass the performance of GPT-4 and ChatGPT significantly.

\begin{table}[htbp]
    \centering
    \renewcommand{\arraystretch}{1.15}
    \setlength{\tabcolsep}{3mm}{
    \begin{tabular}{c|cc}  
    \toprule
     Model     & ACC   & RE \\
    \midrule
    GPT-4         & 18.84\% & -  \\
    ChatGPT       & 10.00\% & - \\
    \hline
    \hline
    \model-10M    & 61.21\% & 97.83\% \\
    \model-100M   & 70.28\% & 99.28\% \\
    \model-500M   & 89.57\% & 99.41\% \\
    \model-2B     & 93.03\% & 99.71\% \\
    \bottomrule
    \end{tabular}} 
    \vspace{1mm}
    \caption{Performance comparison on an arithmetic dataset containing 9,592 test cases between \model and the leading LLMs.}
    \vspace{-0.4cm}
    \label{tab:arithmetic_results}
\end{table}

Additionally, we conduct a performance comparison of arithmetic tasks among different prominent large language models (LLMs) including GPT-4, ChatGPT, text-davinci-003, code-davinci-002, Galactica, LLaMA, OPT, BLOOM, and GLM. For this comparison, we randomly extract a compact arithmetic dataset containing 100 test cases from the larger dataset discussed earlier. The results of this comparison arithmetic performance are presented in Table~\ref{tab:arithmetic_results_1000}. Upon analyzing the results, it is evident that \model achieves a high accuracy of 93.03\% with 2 billion model parameters, surpassing all other LLMs. In addition to leading models like GPT-4 and ChatGPT, the large science model Galactica exhibits better performance in arithmetic tasks. This can be attributed to Galactica's training on a large scientific corpus, enabling it to learn the languages of science and comprehend the intricacies of arithmetic tasks. By leveraging the unique characteristics of this dataset, Galactica is able to enhance its understanding and handling of arithmetic tasks, resulting in improved performance. These findings emphasize the significance of domain-specific training and leveraging specialized datasets to enhance model performance. Besides, a step-by-step solution strategy, which involves decomposing complex arithmetic expressions into individual steps, has proven to be effective in improving arithmetic performance. The outstanding performance of \model shows that the language model coupled with a specialized dataset and the step-by-step solution strategy can achieve remarkable performance in arithmetic tasks.

To comprehensively evaluate the arithmetic performance of \model, we also conduct experiments on a newly-generated arithmetic dataset named MATH 401~\cite{yuan2023well} and the corresponding results are reported in Appendix~\ref{appendix: results_in_math401}.

\begin{small}
    \begin{table}[thbp]
    \centering
    \renewcommand{\arraystretch}{1.15}
    \setlength{\tabcolsep}{2mm}{
    \begin{tabular}{c|cc}  
    \toprule
     Model     & ACC   & RE \\
    \midrule
    GPT-4               & 22.22\% & - \\
    ChatGPT             & 13.25\% & - \\
    text-davinci-003    & 9.79\% & - \\
    text-davinci-002    & 4.08\% & - \\
    Galactica-120b      & 7.97\% & - \\
    Galactica-30b       & 7.02\% & - \\
    LLaMA-65b           & 5.02\% & - \\
    OPT-175B            & 3.83\% & - \\
    BLOOM-176B          & 3.96\% & -  \\
    GLM-130B            & 3.06\% & - \\
    \hline
    \hline
    \model-10M  & 64.29\% & 97.96\% \\
    \model-100M   & 73.47\% & 98.23\% \\
    \model-500M   & 89.80\% & 98.82\% \\
    \model-2B     & 94.90\% & 98.98\% \\
    \bottomrule
    \end{tabular}} 
    \vspace{1mm}
    \caption{Overall performance comparison on various LLMs in term of Accuracy.}
    \vspace{-0.6cm}
    \label{tab:arithmetic_results_1000}
    \end{table}
\end{small}

\vpara{Grouped Results.}
To clearly evaluate the arithmetic ability of \model among different operations, we design a series of extended experiments. Specifically, we design small test datasets comprising 100 test cases to respectively evaluate the arithmetica performance of \model in various arithmetic operations, including addition, subtraction, multiplication, and division. These datasets encompass different data formats, such as integers, decimals, percents, fractions and negative numbers. Here, we compare \model with several well-known chat-type LLMs, such as GPT-4, ChatGPT, ChatGLM, and Bard. The arithmetic performance comparison among these different language models is demonstrated in Table~\ref{tab:group_results}. Analyzing the results, we can observe that the majority of LLMs exhibit commendable accuracy levels exceeding 90\% across diverse data formats for elementary arithmetic operations like addition and subtraction. However, as the complexity escalates to operations like multiplication and division, a divergence in performance manifests across different models. For instance, the accuracy levels of the most powerful model GPT-4 also show a trend towards zero, especially when dealing with decimal and percentile data formats. In contrast, \model consistently shows superior performance in multiplication operations across various data formats, surpassing the capability of GPT-4. This demonstrates the effectiveness and capabilities of \model in handling complex arithmetic tasks, even outperforming a prominent model like GPT-4 in specific operations. Notably, even the smaller variant of \model, \model-10M, with only 10 million training parameters, also achieves remarkable arithmetic performances, further emphasizing the arithmetic capabilities of our \model.

\begin{small}
    \begin{table}[h]
    \centering
    \renewcommand{\arraystretch}{1.15}
    \setlength{\tabcolsep}{1.5mm}{
    \begin{tabular}{c|c|cccc|cc}  
    \toprule
     Task   & Format & GPT-4 & ChatGPT & ChatGLM & Bard & \model-10M & \model-2B\\
    \midrule
    \multirow{5}{*}{ADD} & Int  & 100\%    & 100\%   & 94\%  & 96.0\% & 100\% & 100\%  \\
    & Dec                       & 100\%    & 98\%    & 76\%  & 87\% & 96\%& 100\% \\
    & Frac                      & 43.33\%  & 17.02\% & 32.98\% & 14.2\%  & 60.64\% & 100\% \\
    & Perc                      & 100\%    & 90.0\%  & 1\% & 9.6\% & 100\% & 100\% \\
    & Neg                       & 100\%    & 98\%    & 91\% & 95\% & 100\% & 100\% \\ 
    \hline
    \multirow{5}{*}{SUB} & Int  & 100\%    & 97\%    & 89\% & 91\% & 98\% & 100 \%  \\
    & Dec                       & 100\%    & 94\%    & 82\% & 85\% & 98\% & 100\%  \\
    & Frac                      & 52.48\%  & 18.81\% & 3\% & 24.24\% & 68.32\% & 96.04\%  \\
    & Perc                      & 100\%    & 100\%   & 18\% & 0\% & 99\% & 100\%  \\
    & Neg                       & 100\%    & 97\%    & 44\% & 78\% & 100\% & 100\% \\
    \hline
    \multirow{5}{*}{MUL} & Int  & 9\%      & 4\%     & 1\% & 2\% & 77\% & 84\% \\
    & Dec                       & 0\%      & 0\%     & 0\% & 0\% & 3\% & 33\%  \\
    & Frac                      & 5.63\%   & 2.82\%  & 1.41\% & 1.41\% & 67.61\% & 85.92\%  \\
    & Perc                      & 0\%      & 0\%     & 1\% & 0\% & 81\% & 97\% \\
    & Neg                       & 7\%      & 2\%     & 0\% & 0\% & 76\% & 98\%   \\
    \hline
    \multirow{5}{*}{DIV} & Int  & 92\%     & 91\%    & 24\% & 68\% & 99\% & 100\%   \\
    & Dec                       & 93\%     & 88\%    & 60\% & 60\% & 97\% & 98\%   \\
    & Frac                      & 33.44\%  & 29.69\% & 7.81\% & 1.56\% & 73.44\% & 96.88\%  \\
    & Perc                      & 97\%     & 80\%    & 19\% & 15\% & 88\% & 100\%   \\
    & Neg                       & 97\%     & 90\%    & 50\% & 52\% &  96\% & 100\%  \\
    \bottomrule
    \end{tabular}} 
    \vspace{1mm}
    \caption{Arithmetic comparison between \model and other LLMs among different operations. Int denotes integers, Dec denotes decimals, Frac denotes fractions, Perc denotes percents, and Neg denotes negative numbers.}
    \vspace{-0.5cm}
    \label{tab:group_results}
    \end{table}
\end{small}

\vpara{Results in BIG-bench.} We also evaluate \model using BIG-bench arithmetic dataset~\cite{srivastava2022beyond}, which is commonly used to evaluate basic arithmetic capabilities of language models by performing n-digit addition (ADD), subtraction (SUB), multiplication (MUL), and division (DIV). Table~\ref{tab:big-bench-math} reports the experimental results of GPT-4 and \model on various arithmetic operations with different numbers of digits. GPT-4 exhibits near-perfect (100\%) accuracy in low-digit arithmetic tasks. However, as the digits escalate, the performance gradually diminishes, particularly pronounced in the multiplication task. In contrast, \model consistently maintains high accuracy levels even in high-digit arithmetic tasks, illustrating its outstanding ability to handle complex arithmetic tasks effectively. The performance trends of different \model variants reveal a consistent pattern of improvement as model size increases. For ADD and SUB tasks, the accuracy remains consistently high across all model sizes with slight variations. There is a tendency for larger models to achieve higher accuracy compared to smaller models but the differences in performance between different model sizes are relatively small. In the MUL task, accuracy rises distinctly with larger model sizes. Smaller models exhibit relatively lower accuracy, while larger counterparts demonstrate enhanced accuracy, particularly in tasks involving higher digit numbers. A similar tendency can be observed in the DIV task. Overall, the evaluation results demonstrate that \model outperforms GPT-4 in high-digit arithmetic tasks, and the performance generally inproves with larger model sizes.

\begin{small}
    \begin{table*}[h]
    \centering
    \renewcommand{\arraystretch}{1.15}
    \setlength{\tabcolsep}{2mm}{
    \begin{tabular}{c|c|c|cccc}  
    \toprule
     Task   &  &GPT-4 & \model-10M   & \model-100M  & \model-500M & \model-2B \\
    \midrule
    \multirow{5}{*}{ADD} & 1D  & 100\% & 84\% & 100\% & 100\% & 100\% \\
    & 2D & 100\% & 97.2\% & 100\% & 100\% & 100\% \\
    & 3D & 99.6\%  & 99.3\% & 100\% & 100\%  & 100\% \\
    & 4D & 98.8\%  & 99.9\% & 99.9\% & 100\% & 100\% \\
    & 5D & 94.1\% & 99.2\% & 100\% & 99.6\% & 99.4\%  \\ 
    \hline
    \multirow{5}{*}{SUB} & 1D & 100\% & 92\% & 100\% & 100\% & 100\%  \\
    & 2D & 100\%  & 98.5\%    & 99.8\% & 100\% & 100\% \\
    & 3D & 99.2\% & 98.8\%  & 99.9\% & 100\% & 99.9\% \\
    & 4D & 98.9\% & 98.4\%  & 99.6\% & 99.7\% & 99.8\% \\
    & 5D & 92.4\% & 98.0\%  & 99.3\% & 99.5\%  & 98.9\% \\
    \hline
    \multirow{5}{*}{MUL} & 1D & 100\% & 91\% & 100\% & 99\%  & 100\% \\
    & 2D & 99.4\% & 85.8\% & 99.7\%  & 99.9\% & 99.9\% \\
    & 3D & 30.3\% & 77.8\% & 91.4\% & 93.7\% & 98.3\% \\
    & 4D & 5.3\%  & 79.7\% & 80.4\% & 90.0\%  &  94.9\% \\
    & 5D & 0.0\% & 41.6\% & 55.6\%  & 59.6\% &  89.9\% \\
    \hline
    \multirow{5}{*}{DIV} & 1D & 100\%& 87.0\% & 100\% & 100\% &100\%  \\
    & 2D & 100\% & 89.5\% & 100\%  & 100\% & 100\% \\
    & 3D & 94.5\% & 90.2\% & 100\% & 99.6\%  & 99.4\% \\
    & 4D & 90.9\%  & 90.5\% & 99.5\%  & 99.6\%  & 100\% \\
    & 5D & 53.4\%  & 82.2\% & 92.9\%  & 93.6\% & 94.9\% \\
    \bottomrule
    \end{tabular}} 
    \caption{Overall performance comparison on GPT-4 and \model on BIG-bench Arithmetic sub-task.}
    \vspace{-0.7cm}
    \label{tab:big-bench-math}
    \end{table*}
\end{small}

\vpara{Analysis on \model.} Despite achieving an impressive overall accuracy of $93.03\%$ with its 2 billion model parameters, a thorough analysis is conducted to comprehend instances where \model fails to generate accurate answers. Consider the example $3468*4046/7424$, \model generate an answer of $468*4046/7424=14031528/7424=1889.901400862069$, while the true answer is $468*4046/7424=14031528/7424=1890.0226293103$. Upon comparing the generated results with the true answers, it is obviously observed that the multiplication operation for $468*4046$ is correct but the division operation for $14031528/7424$ is incorrect. One possible reason for this discrepancy is that \model's pre-training primarily encompasses numbers in the 5-digit range, thereby causing inaccuracies when tackling division tasks involving 12-digit and 4-digit numbers. Upon thorough analysis of the errors made by \model, it's important to highlight that the inaccuracies in the generated answers are remarkably close to the correct evaluations. For a comprehensive investigation into the errors, a detailed breakdown of the error types and their frequencies can be found in Appendix~\ref{appendix: cases on char_error}.

\subsubsection{Ablation Study}

\begin{figure*}[htbp]
    \centering
    \includegraphics[width=\textwidth]{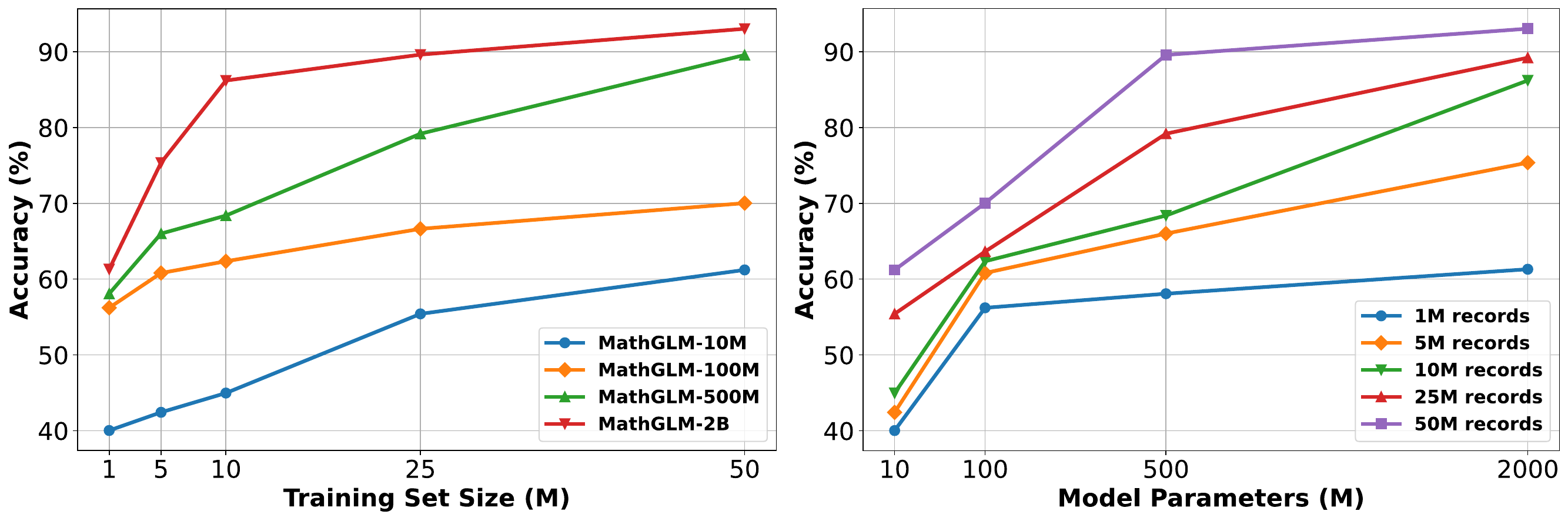}
    \caption{Performance visualization on \model under different scaling configurations, including model parameters and training data sizes.
    }
    \vspace{-0.4cm}
    \label{fig:scale_analysis}
\end{figure*}

\vpara{Scaling Analysis.} To comprehensively assess the effect of model parameters and training data sizes on performance, we conduct a series of scaling analysis experiments. The model parameters of \model are designed as a range of $\{10M, 100M, 500M, 2B\}$ and the training data sizes is set to a range of $\{1M, 5M, 10M, 25M, 50M\}$. Figure~\ref{fig:scale_analysis} shows the evaluation performance of \model under various scaling configurations. As expected, the performance trend highlights that the 2B model consistently outperforms its smaller counterparts when evaluated using equivalent data sizes, illustrating the positive impact of larger model parameters on arithmetic performance. Besides, it is evident that larger data sizes have a substantial influence on improving the arithmetic performance as well. However, it is important to note that the effect of data size on the smaller model sizes may not be as pronounced as compared to the larger models. This discernible pattern implies that the benefits derived from increasing the data size tend to be more substantial when paired with larger model parameters. In essence, the trends illustrated in Figure~\ref{fig:scale_analysis} substantiate the notion that both the size of the model and the quantity of training data play vital roles in enhancing the arithmetic performance of \model.

\begin{figure*}[htbp]
    \centering
    \includegraphics[width=0.8\textwidth]{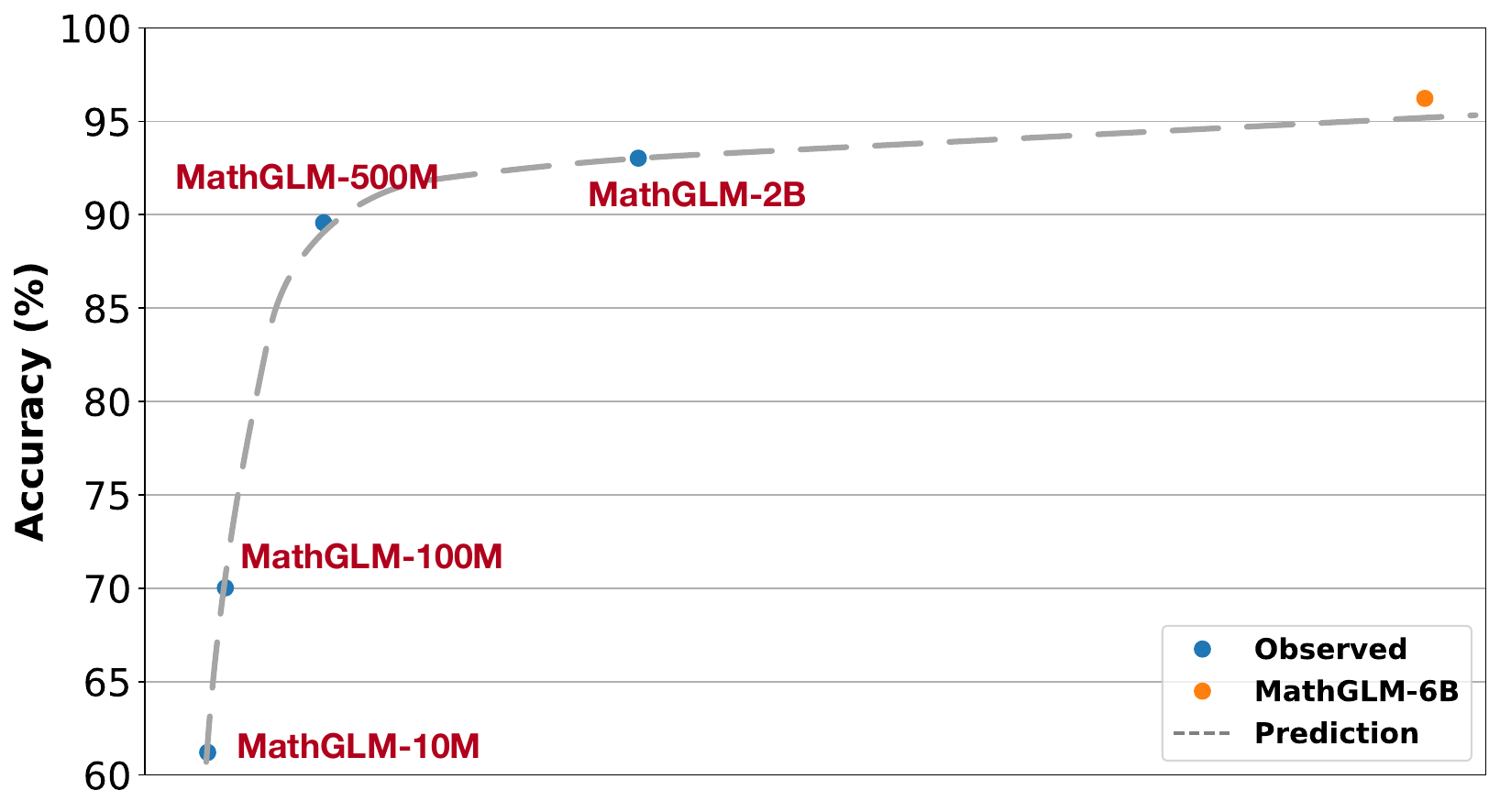}
    \caption{The log-linear trend exhibited by the \model. This trend accurately predicts \model-6B's performance. 
    }
    \vspace{-0.5cm}
    \label{fig:prediction}
\end{figure*}

Furthermore, by analyzing the trend illustrated in Figure~\ref{fig:scale_analysis}, we attempt to extend our findings and make predictions for scaling configurations that were not directly studied. Employing a log-linear trend assumption, we can extrapolate the results to estimate the requisite model size for achieving a targeted performance when utilizing a more extensive training set. Figure~\ref{fig:prediction} illustrates the extrapolated outcomes derived from the log-linear trend. To validate the validity of this trend, we pre-train a \model equipped with 6B model parameters. From Figure~\ref{fig:prediction}, we can observe that the extrapolated trend aligns with the performance achieved by the \model-6B.

\vpara{Generalization Analysis.} To assess the generalization ability of \model beyond the 5-digit range, a set of 50,000 training records involving numbers within the 12-digit range are introduced into the training dataset. After incorporating this additional data, \model is further pre-trained for 20,000 steps to enhance its ability to handle arithmetic tasks involving numbers outside the 5-digit range. Table~\ref{tab:model_generalization} shows the arithmetic performance comparison across various digit ranges, spanning from 5 digit to 12 digit, and involving a mix of arithmetic operations.  In comparison to GPT-4 and ChatGPT, our proposed \model consistently achieves the highest accuracy across all digit ranges, indicating the superiority of \model for multi-digit arithmetic operations. A noticeable observation is that a decline in accuracy as the number of digits in the arithmetic operations increases. This suggests that handling larger digit ranges poses a greater challenge to all LLMs.

\begin{small}
\begin{table}[hbpt]
    \centering
    \renewcommand{\arraystretch}{1.15}
    \setlength{\tabcolsep}{1mm}{
    \begin{tabular}{c|cc|cc}  
    \toprule
     Generalization   & GPT4 & ChatGPT & \model-500M & \model-2B \\
    \midrule
    5-digit    & 6.67\% & 5.43\% & 83.44\% & 85.16\% \\ 
    6-digit    & 10.0\% & 2.94\% & 79.58\% & 78.17\% \\ 
    7-digit    & 3.33\% & 1.92\% & 71.19\% & 73.73\%  \\ 
    8-digit    & 3.13\% & 1.43\% & 64.62\% & 67.69\%  \\ 
    9-digit    & 6.90\% & 1.57\% & 66.66\% & 69.60\% \\ 
    10-digit   & 3.33\% & 1.45\% & 49.55\% & 65.77\% \\ 
    11-digit   & 0\% & 0\% & 42.98\% & 57.89\%  \\ 
    12-digit   & 6.90\% & 1.33\% & 27.38\% & 41.05\% \\ 
    \bottomrule
    \end{tabular}}
    \vspace{1mm}
    \caption{Performance comparison between most powerful LLMs and \model on various multi-digit arithmetic operations.
    }
    \vspace{-0.6cm}
    \label{tab:model_generalization}
\end{table}
\end{small}

\vpara{Step-by-step Analysis.} To delve deeper into the impact of the step-by-step strategy on \model, we conduct extended experiments that directly calculate the answer of each arithmetic expression without employing the step-by-step approach. Figure~\ref{fig:step} shows performance comparison between employing the step-by-step strategy and bypassing it for different models. We can observe that a significant improvement in the peformance of \model when the step-by-step strategy is applied. For instance, in the case of \model-500M, the accuracy rises from 31.96\% to 89.57\%, while for \model-2B, it increases from 40.76\% to 93.03\% for \model-2B, all attributable to the incorporation of the step-by-step strategy. Similarly, the relative error accuracy exhibits a similar positive trend, escalating from 89.29\% to an exceptional 99.41\% for \model-500M, and from 94.26\% to an outstanding 99.71\% for \model-2B with the implementation of the step-by-step strategy. These results demonstrate the effectiveness of the step-by-step strategy in enhancing \model's ability to accurately perform arithmetic operations. The step-by-step approach enables \model to better understand and solve intricate arithmetic tasks, leading to significant improvements in accuracy and relative error accuracy metrics.

\begin{figure*}[htbp]
    \centering
    \includegraphics[width=0.8\textwidth]{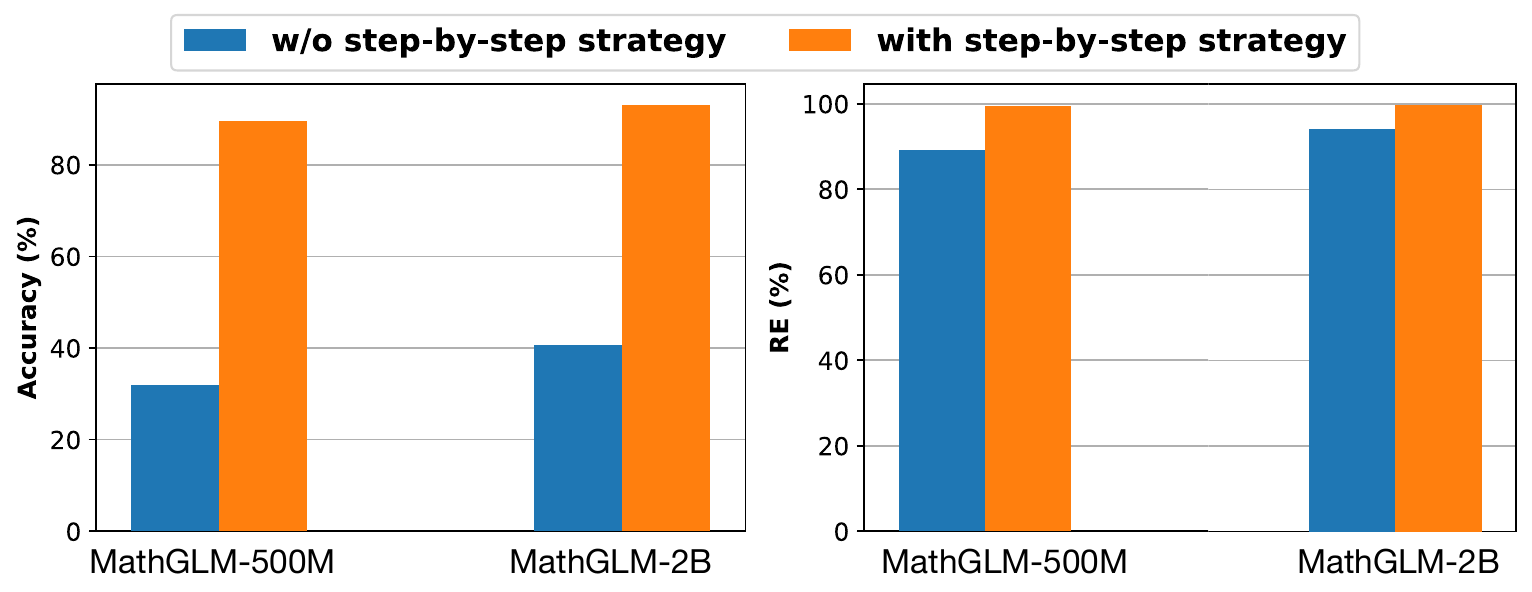}
    \caption{Performance comparison of \model with and without the step-by-step solution.
    }
    \vspace{-0.3cm}
    \label{fig:step}
\end{figure*}

\subsection{Learning on Math Word Problems}\label{subsec:exp_mwp}

\subsubsection{Dataset}

In the field of math word problems (MWP), the performance of \model is measured using the Ape210K dataset~\cite{zhao2020ape210k}, which contains a collection of 5,000 test math problems. Additionally, we introduce the K6 dataset, which is designed to cover  math word problems suitable for elementary school students across 6 different grade levels. The primary purpose of the K6 dataset is to assess the mathematical abilities of LLMs in comprehending and solving general-purpose math reasoning problems. 
By evaluating \model on the K6 dataset, we are able to gauge its effectiveness in handling mathematical word problems of varying complexity and across a range of grade levels. The details of the K6 dataset can be found in Appendix~\ref{appendix: k12 dataset}.

\subsubsection{Overall Results}
To assess the effectiveness of \model, we test it on the Ape210K dataset and a newly-collected K6 dataset. To facilitate these evaluations, we utilize various LLMs as the backbone. These LLMs, including GLM-Large, GLM-6B, GLM2-6B, GLM-10B, ChatGLM-6B, and ChatGLM2-6B, are employed as the core architecture to process and comprehend mathematical word problems within Chinese datasets.

\vpara{Results on the Ape210K dataset.} We report the performance results of various LLMs including GPT-4, ChatGPT, and a series of our \model variations in Table~\ref{tab:mwp_results_ape210k}. The results show that when paired with GLM-10B, \model achieves performance levels comparable to the state-of-the-art GPT-4 model in terms of answer accuracy. It demonstrates the effectiveness of \model in generating accurate answers for math word problems through the utilization of a step-by-step strategy. Furthermore, we report the arithmetic accuracy, which measures the correctness of the generated arithmetic expressions. Notably, \model consistently achieves higher arithmetic accuracy compared to answer accuracy across different model sizes. A distinct trend emerges when comparing \model's performance with GLM-Large, GLM-6B, and GLM-10B: \model exhibits notable enhancements in both arithmetic accuracy and answer accuracy. This observation indicates that augmenting model size tends to bolster its overall performance. However, it is worth noting that the performance of \model drops significantly compared to the GLM models when it is coupled with ChatGLM models. A possible explanation is that ChatGLM models are fine-tuned using the instruction data, potentially compromising the inherent capabilities of language models. This tuning process might introduce biases or constraints that hinder the overall ability of the language models in handling math word problems.

\begin{table}[hbpt]
    \centering
    \renewcommand{\arraystretch}{1.15}
    \setlength{\tabcolsep}{1mm}{
    \begin{tabular}{c|cc}  
    \toprule
     Model    & $\text{Arithmetic}_{Acc}$ & $\text{Answer}_{Acc}$   \\
    \midrule
    GPT-4 & - & 59.57\%  \\
    GPT-3.5-turbo & -  & 39.78\% \\
    \hline 
    \hline 
    GLM-Large & - & 0\%\\
    + MathGLM &  62.00\%  & 50.80\%   \\
    \hline
     GLM-6B &  - & 3.94\%  \\
    + MathGLM & 64.60\%   & 48.06\%   \\
    \hline
    GLM-10B & - & 0\% \\
    + MathGLM &  69.08\%  & 58.68\%  \\
    \hline
    \hline
     GLM2-6B &  -  &  31.42\% \\
    + MathGLM &  52.24\% & 45.48\%   \\
    \hline
    \hline
    ChatGLM-6B &  - & 6\%  \\
    + MathGLM &  58.52\% & 42.28\%  \\
    \hline
    ChatGLM2-6B &  -  & 31.70\% \\
    + MathGLM & 50.38\%   & 43.14\%    \\
    \bottomrule
    \end{tabular}} 
    \vspace{1mm}
    \caption{Performance comparison among different language models on the Ape210K dataset.
    }
    \vspace{-0.5cm}
    \label{tab:mwp_results_ape210k}
\end{table}

\begin{figure*}[htbp]
    \centering
    \includegraphics[width=0.98\textwidth]{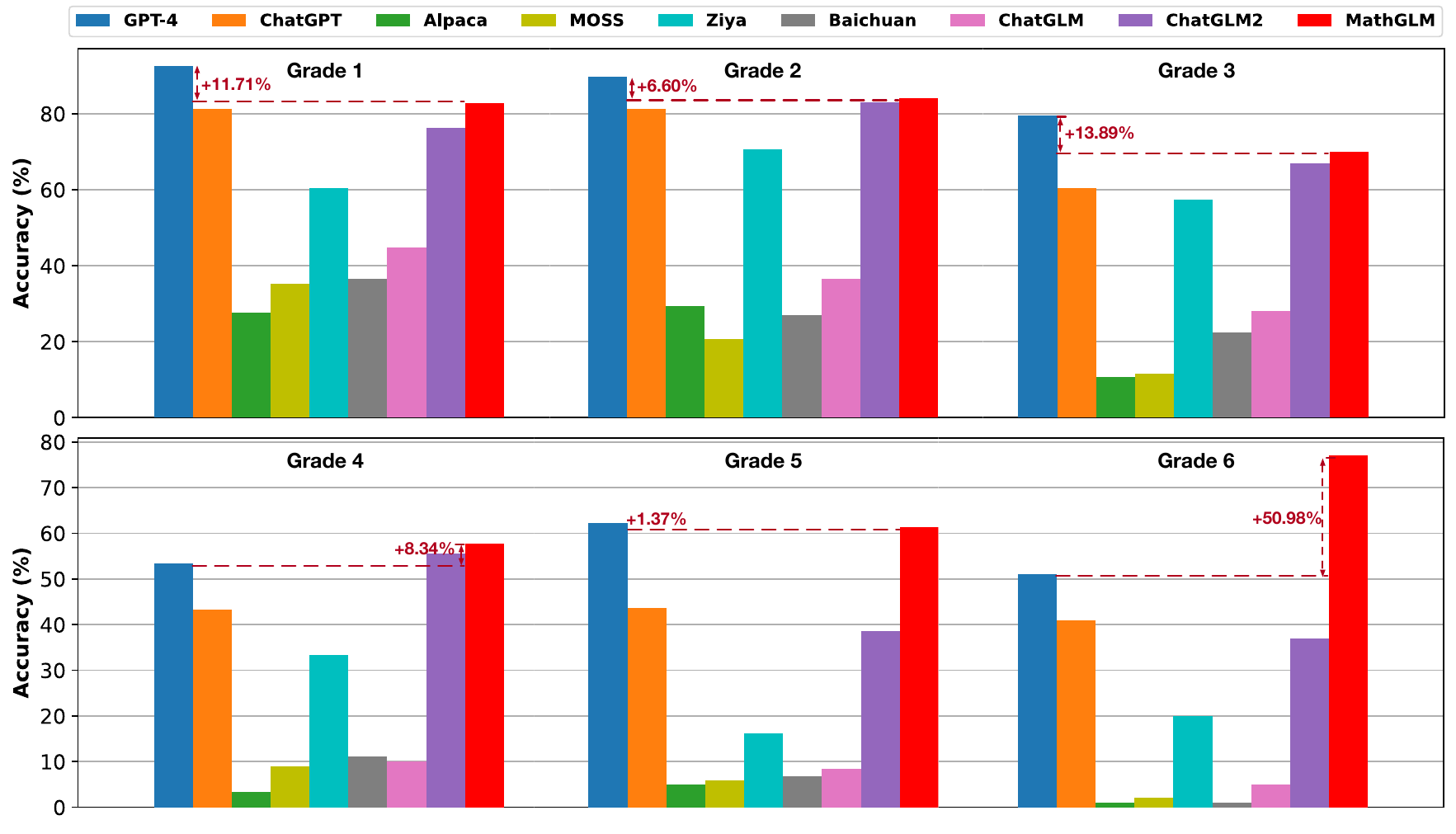}
    \caption{Performance comparison between \model and other popular language models on the K6 dataset.
    }
    \vspace{-0.3cm}
    \label{fig:k12_results}
\end{figure*}

\vpara{Results on the K6 dataset.} To assess the mathematical problem-solving abilities across different grade levels, we introduce the K6 dataset and present the corresponding performance results for various LLMs in Figure~\ref{fig:k12_results}. The figure shows the overall performance results for GPT-4, ChatGPT, Chinese-Alpaca-13B, MOSS-16B, Ziya-LLaMA-13B, Baichuan-7B, ChatGLM-6B, ChatGLM2-6B, and \model-GLM-10B across each individual grade level. The detailed introduction of these models is provided in Appendix~\ref{appendix:baseline_models}. 
The observations from the figure indicate a general trend of performance decreases as the grade level increases. Such observation indicates that solving math word problems becomes progressively more challenging for LLMs as the grade level increases, requiring more advanced problem solving skills and a deeper understanding of mathematical concepts.
GPT-4 exhibits consistently high accuracy levels across most grade levels, showcasing its proficiency in handling math word problems spanning various educational stages. Comparatively, ChatGPT outperforms the majority of Chinese LLMs in terms of accuracy across different grade levels. Among the evaluated Chinese LLMs, ChatGLM2-6B demonstrates a commendable level of performance, achieving satisfactory accuracy (reaching 60\% accuracy) in solving math word problems from grade 1 to 4. However, its effectiveness diminishes when attempting to solve problems in grade 5 and 6, highlighting challenges in handling more complex problem-solving scenarios at those levels. 
\model consistently outperforms ChatGPT and many of the most powerful Chinese Language Models (LLMs) across the spectrum of grade levels, from grade 1 to grade 6. Particularly noteworthy is \model's ability to achieve higher accuracy than GPT-4 in more advanced grades, such as grade 5 and 6. This observations show the effectiveness of \model in enhancing the accuracy of solving math word problems, especially in challenging educational contexts that demand deeper mathematical understanding and advanced problem-solving skills.

\subsubsection{Comparison of Training Strategies} 

Here, we evaluate the mathematical reasoning ability of \model with different training strategies: fine-tuning and continue training. To execute continue training, we amalgamate the Ape210K train dataset with instruction data released by Chinese-Vicuna~\cite{leng2023chinese-vicuna}. We subsequently continue training \model from the GLM-10B backbone. Table~\ref{tab:generic_perf} shows the overall performance comparison of \model employing different training strategies. We observe that directly fine-tuning on the specific dataset can achieves better performance.

\begin{small}
    \begin{table*}[h]
    \centering
    \renewcommand{\arraystretch}{1.15}
    \setlength{\tabcolsep}{2mm}{
    \begin{tabular}{c|cc|cc}  
    \toprule
    \multirow{2}{*}{Training}     & \multicolumn{2}{c|}{w/o step-by-step strategy }   & \multicolumn{2}{c}{with step-by-step strategy}  \\
     &  $\text{Arithmetic}_{Acc}$ & $\text{Answer}_{Acc}$ & $\text{Arithmetic}_{Acc}$ & $\text{Answer}_{Acc}$ \\
    \midrule
    Fine-tuning & 71.38\%  & 41.24\%  & 69.08 \% & 58.68\% \\
    \hline
    Continue training & 70.16\% & 40.34\% & 67.02\% & 56.60\%  \\
    \bottomrule
    \end{tabular}} 
    \caption{Overall performance comparison on various LLMs in term of Accuracy.}
    \vspace{-0.5cm}
    \label{tab:generic_perf}
    \end{table*}
\end{small}

\subsubsection{Further Analysis}

\vpara{Scaling Analysis.} To explore the impact of scaling on \model, we conduct a series of experiments encompassing varying dataset sizes and distinct model parameters. Table~\ref{tab:mwp_scale_zh} demonstrates the results obtained from varying the dataset sizes within the range of $\{5K, 10K, 20K, 50K, 100K, 200K\}$.  Furthermore, to understand the impact of different model parameters, we incorporate various backbone models into \model, including GLM-Large (335M), GLM-6B, and GLM-10B. The results consistently indicate that \model's performance improves across all  backbone models with the increase in dataset size. Such observation highlights the beneficial effects of enlarging the training data on bolstering \model's proficiency in tackling math word problems. By accessing more extensive datasets, \model is introduced to a wider array of problem types, resulting in better performance. Additionally, discernible differences in performance emerge among the various backbone models. Given sufficient dataset size, larger models like \model-GLM-10B often outperform others, indicating the crucial role of model parameters in addressing intricate math word problems. These insights emphasize the significance of both dataset and model scaling. By augmenting dataset size and utilizing larger models, we can markedly boost \model's capability to generate more accurate solutions, enhancing its overall efficacy in resolving math word problems.

\begin{table*}[hbpt]
    \centering
    \renewcommand{\arraystretch}{1.15}
    \setlength{\tabcolsep}{2mm}{
    \begin{tabular}{c|ccc}  
    \toprule
     Model Scale    & \model-GLM-Large   &\model-GLM-6B  &  \model-GLM-10B   \\
    \midrule
    5K Problems      & 4.32\%   & 12.84\% &  3.68\%    \\
    10K Problems     & 7.14\%   & 19.78\% &  6.36\%    \\
    20K Problems     & 10.36\%  & 21.89\%  &  9.62\%      \\
    50K Problems     & 18.32\%  & 26.40\%  &  16.78\%    \\
    100K Problems    & 25.98\%  & 31.44\%  &  22.20\%     \\
    200K Problems    & 35.68\%  & 34.00\% &  38.10\%     \\
    \bottomrule
    \end{tabular}} 
    \caption{Performance comparison of \model on different training dataset sizes and model parameters.
    }
    \vspace{-0.3cm}
    \label{tab:mwp_scale_zh}
\end{table*}

\begin{figure*}[htbp]
    \centering
    \includegraphics[width=\textwidth]{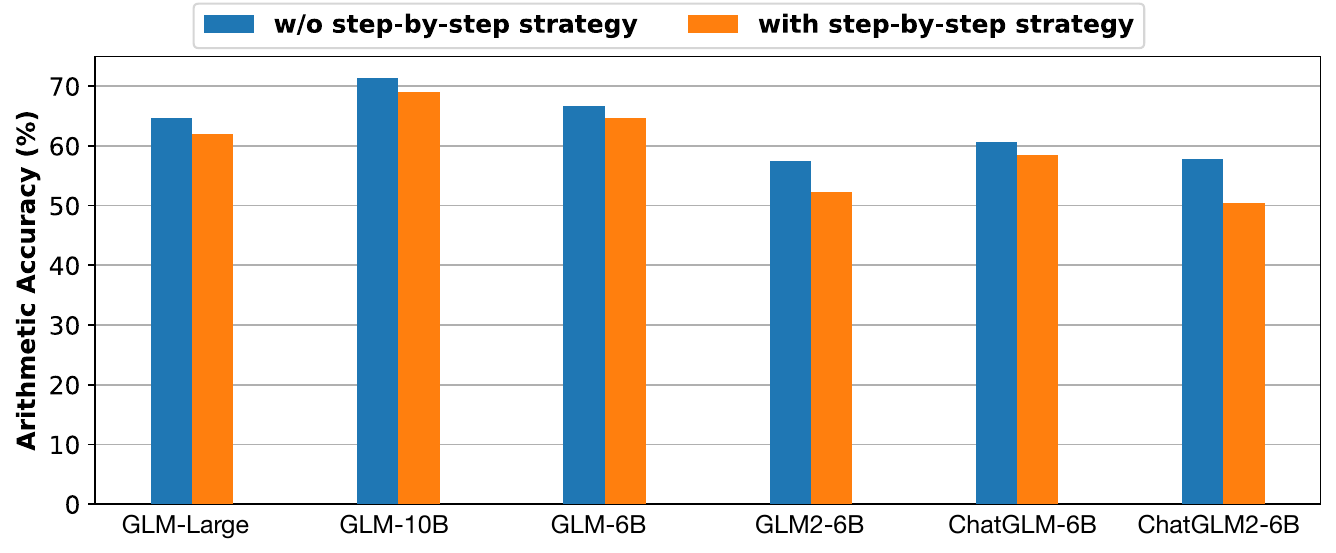}
    \caption{The arithmetic accuracy of \model is evaluated across various backbone models, considering both with and without the implementation of a step-by-step strategy. Interestingly, there's a slight decrease in arithmetic accuracy when the step-by-step method is employed, likely due to the model having to perform calculations sequentially for each math problem.
    }
    \vspace{-0.3cm}
    \label{fig:step_arithmetic}
\end{figure*}

\vpara{Step-by-Step Analysis for MWP.} To investigate the impact of the step-by-step strategy on \model, we conduct a series of ablation studies to explore the performance differences with and without this strategy. Figure~\ref{fig:step} and Figure~\ref{fig:step_arithmetic} demonstrate the performance comparison of \model across different GLM and ChatGLM models respectively. In terms of arithmetic accuracy, as shown in Figure~\ref{fig:step_arithmetic}, the \model equipped with the step-by-step strategy records marginally lower scores than its counterpart without the strategy. This can be attributed to the fact that the step-by-step approach necessitates a sequential calculation for each mathematical problem. This encourages \model to concentrate on grasping the foundational mathematical rules. Consequently, a portion of the \model's processing power is dedicated to understanding and generating step-by-step solutions, which might slightly weaken its prowess in precisely crafting arithmetic expressions. Nevertheless, while there's a minor dip in arithmetic accuracy, the step-by-step strategy significantly bolsters \model's answer accuracy. By guiding \model to derive answers progressively, this approach ensures \model generates higher accuracy in solving math word problems. Notably, we observe pronounced improvements in answer accuracy across all GLM variants: 37.86\% for GLM-Large, 42.29\% for GLM-10B, 47.97\% for GLM-6B, and 53.96\% for GLM2-6B. Similar trends are also evident in the ChatGLM models, recording gains of 40.65\% in ChatGLM-6B and 49.38\% in ChatGLM2-6B. These results highlight the inherent trade-off between arithmetic accuracy and answer accuracy by employing the step-by-step strategy. While this strategy may introduce some potentially impact on arithmetic accuracy, it effectively enhance \model's ability to generate accurate answers for math word problems.

\begin{figure*}[htbp]
    \centering
    \includegraphics[width=\textwidth]{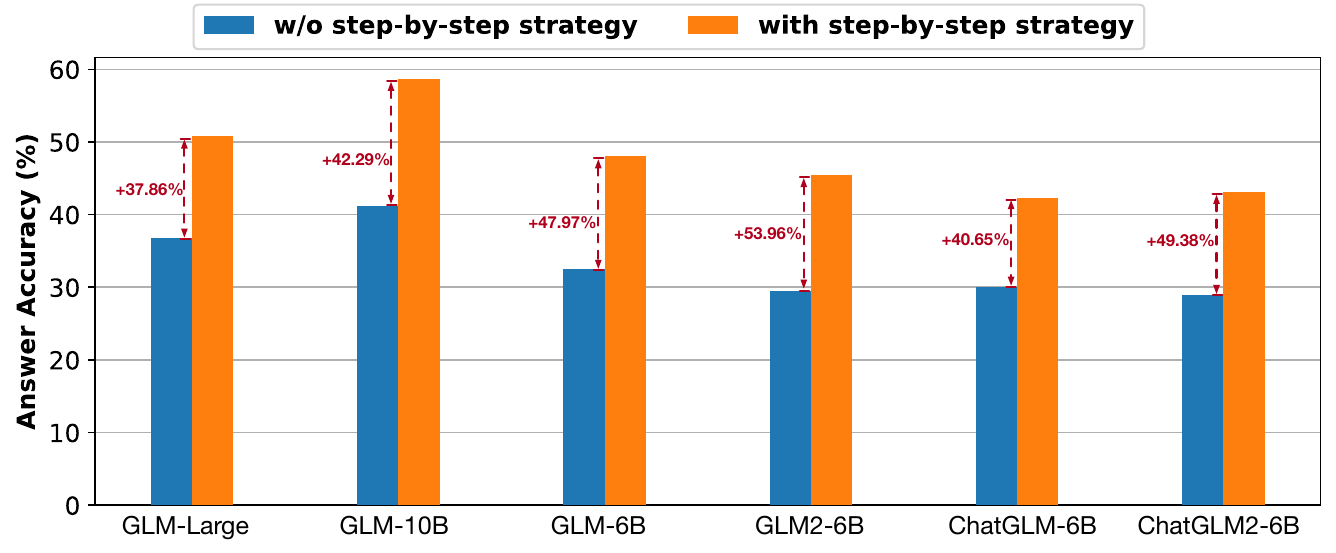}
    \caption{The answer accuracy of \model is compared across various backbone models, both with and without the use of a step-by-step strategy. Employing the step-by-step approach, we observe a marked improvement in answer accuracy relative to the model's performance without it.
    }
    \label{fig:step}
\end{figure*}

\subsubsection{Failure Analysis on Math Word Problems}\label{appendix: mwp_fail_analysis}

\begin{figure}[htbp]
    \centering
    \includegraphics[width=\textwidth]{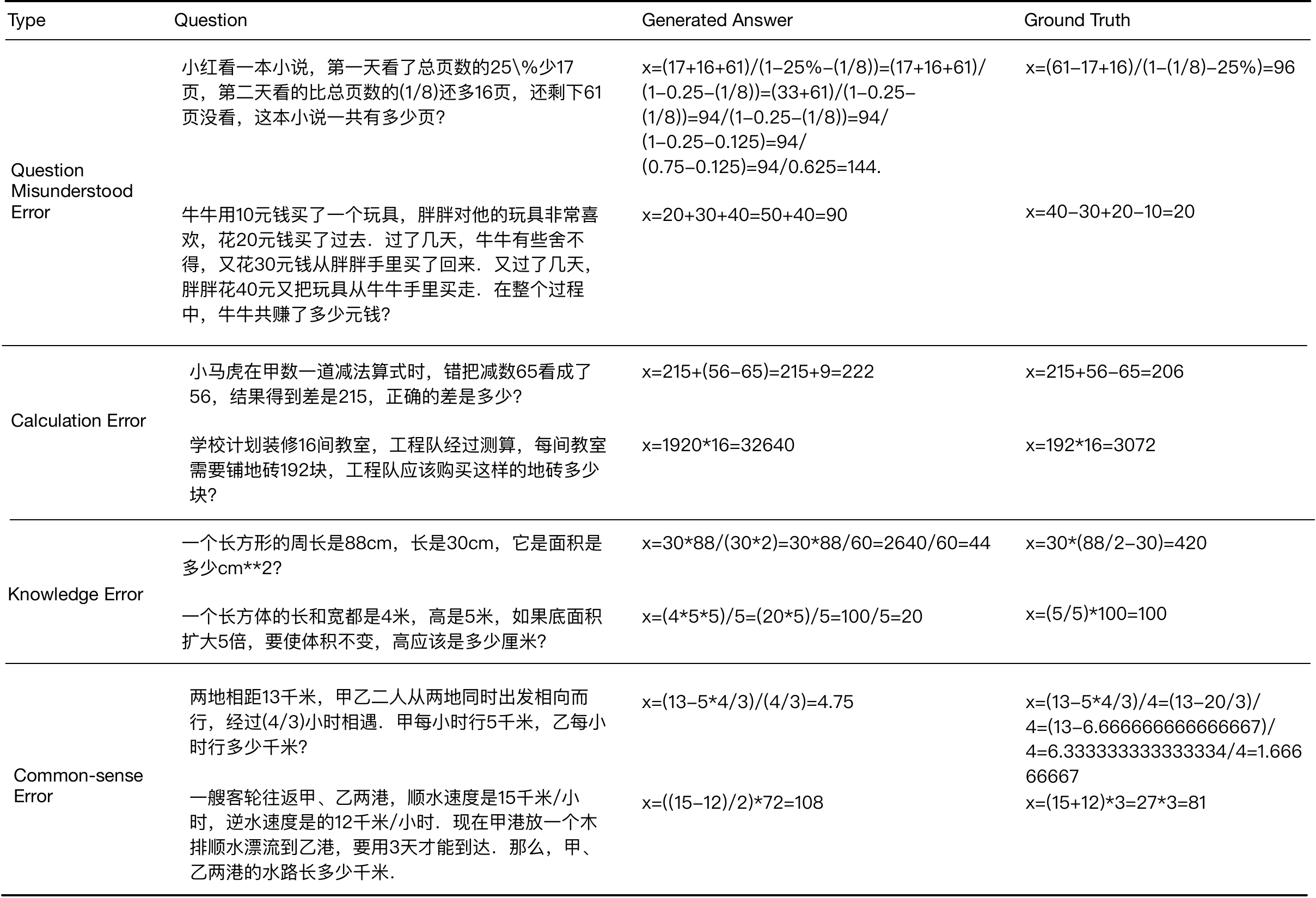}
    \caption{Some failed examples generated by \model-GLM-10B on solving math word problems.
    }
    \vspace{-0.5cm}
    \label{fig:fails_mwp}
\end{figure}

Figure~\ref{fig:fails_mwp} provides some failed examples generated by \model-GLM-10B on solving math word problems. We can identify certain challenging scenarios where \model-GLM-10B encounters difficulties in solving math word problems. One common issue is the misinterpretation of ambiguous language, leading to incorrect problem-solving approaches. For instance, ambiguous phrases such as ``more than'' or ``less than'' can be interpreted differently by the model, resulting in inaccurate solutions. Additionally, \model-GLM-10B tends to struggle with problems that involve complex mathematical operations. As a result, it may provide partially correct arithmetic solutions but fail to arrive at the final correct answer. 

\begin{figure}[htbp]
    \centering
    \includegraphics[width=0.8\textwidth]{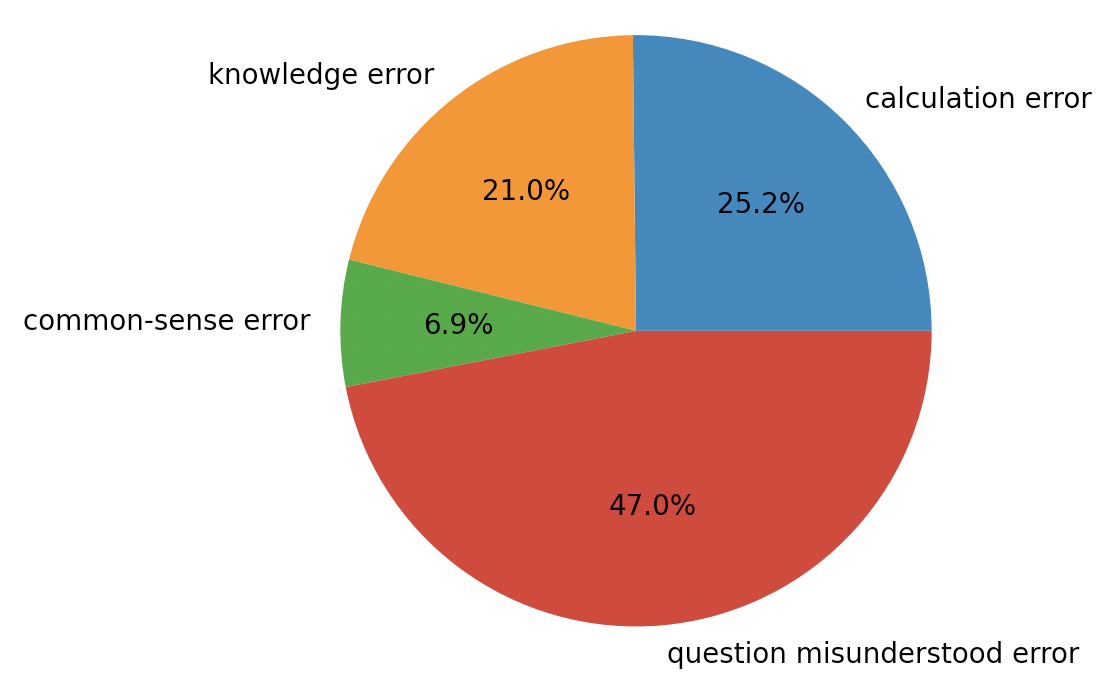}
    \caption{The distribution of error types generated by \model-GLM-10B on math word problems.
    }
    \vspace{-0.3cm}
    \label{fig:mwp_error_dis}
\end{figure}

Here, we construct a percentile graph to analyze the distribution of error types made by the \model-GLM-10B on the Ape210K test dataset. As shown in Figure~\ref{fig:mwp_error_dis}, we can identify the most common error types that may require improvement for the \model-GLM-10B. One prominent error type that stands out is question misunderstood errors. These errors occur when the \model-GLM-10B misunderstands the language and context of certain math word problems, leading to inaccurate problem-solving solutions.
Despite these limitations, it is worth noting that \model-GLM-10B still demonstrates a remarkable ability to solve a wide range of math word problems accurately.

\section{Conclusion}\label{sec:conclusion}

In this paper, our primary focus revolves around evaluating the mathematical reasoning capabilities of LLMs, encompassing both arithmetic operations and math word problems. For arithmetic tasks, we incorporate step-by-step solution and curriculum learning to train a Transformer-based language model from scratch. With comprehensive training on ample data, we establish that a language model boasting 2 billion parameters can achieve outstanding accuracy in multi-digit arithmetic tasks, exceeding GPT-4's results by a considerable margin. This finding compellingly challenges the prevailing cognition that LLMs face constraints in executing accurate arithmetic operations, especially when dealing with multi-digit numbers, decimals, and fractions, without leaning on external computational aids. When pivoting to math word problems, we reconstruct a dataset enriched with multi-step arithmetic operations. After fine-tuning our \model on this revamped dataset derived from GLM-10B, it achieves similar performance to GPT-4 on the 5,000-sample test set of Chinese math problems, demonstrating its formidable prowess.

{\small
\bibliographystyle{plainnat}
\bibliography{reference}
}

\newpage
\appendix
\section{Appendix}

\subsection{Tokenization for Arithmetic Tasks}\label{appendix:arithmetical_hyperparameters}

The arithmetic operations in our \model involve numbers from 0 to 9, and the calculating signs comprise addition (+), subtraction (-), multiplication (*), division (/), and exponentiation ($\textasciicircum$). Symbols that represent forms in the data include the decimal point (.), percent sign (\%), negative sign (-), fraction delimiter (/), brackets such as '(' and '[', and the equal sign (=).

To achieve a consistent tokenization process, we adopt the unified tokenization tool \textit{icetk} proposed in CogView2~\cite{ding2022cogview2}. By leveraging this methodology, we tokenize each digit as a distinct token. For instance, the numeral ``12345'' is tokenized into the set $\{1,2,3,4,5\}$. To allocate singular tokens to the other mentioned symbols, we disengage the continuous representation symbols within icetk throughout the tokenization procedure. 

Table~\ref{tab:tokenization} shows some tokenization examples employed in \model. This tokenization approach ensuers that every element in the arithmetic expression is adequately represented and can be efficiently processed by the \model, facilitating \model to excute comprehensive arithmetic tasks. 

Owing to the variable lengths of arithmetic expressions, it becomes imperative to standardize their lengths for efficient training of the \model. A straightforward method, like padding each input to a fixed length, might damage training efficacy. To circumvent this, we adopt a more efficient strategy, where multiple arithmetic expressions are concatenated until they achieve a predefined fixed length.

\begin{small}
\begin{table*}[hbpt]
    \centering
    \renewcommand{\arraystretch}{1.25}
    \setlength{\tabcolsep}{0.1mm}{
    \small
    \begin{tabular}{p{3.5cm}p{10cm}}  
    \toprule
     Input &  Tokenization    \\
    \midrule 
    \multirow{3}{*}{12345+345=} & ['\_', '1', '2', '3', '4', '5', '+', '3', '4', '5', '=']   \\
    ~ & [20005, 20009, 20010, 20013, 20016, 20015, 20065, 20013, 20016, 20015, 20054] \\
     \hline
     \multirow{3}{*}{1234-45678=} & ['\_', '1', '2', '3', '4', '-', '4', '5', '6', '7', '8', '='] \\
     ~ & [20005, 20009, 20010, 20013, 20016, 20011, 20016, 20015, 20021, 20025, 20023, 20054] \\
     \hline
     \multirow{2}{*}{34*678=} & ['\_', '3', '4', '*', '6', '7', '8', '='] \\
     ~ & [20005, 20013, 20016, 20032, 20021, 20025, 20023, 20054]\\
      \hline
     \multirow{2}{*}{1.2/2=} & ['\_', '1', '.', '2', '/', '2', '='] \\
     ~ & [20005, 20009, 20007, 20010, 20026, 20010, 20054] \\
      \hline
     \multirow{5}{*}{(1.2*3\%)/2+[(12+3)*5]=} & ['\_', '(', '1', '.', '2', '*', '3', '\%', ')', '/', '2', '+', '[', '(', '1', '2', '+', '3', ')', '*', '5', ']', '='] \\
     ~ & [20005, 20020, 20009, 20007, 20010, 20032, 20013, 20040, 20014, 20026, 20010, 20065, 20052, 20020, 20009, 20010, 20065, 20013, 20014, 20032, 20015, 20042, 20054]\\
    \bottomrule
    \end{tabular}} 
    \caption{Some examples of tokenization in \model.}
    \vspace{-0.3cm}
    \label{tab:tokenization}
\end{table*}
\end{small}

\subsection{Backbone Models}\label{appendix:backbone_models}

General Language Model (GLM) is a Transformer-based language model that combines autogressive blank infilling with bidirectional attention mechanisms. Different from decoder-only language models that primarily rely on unidirectional attention, GLM integrates bidirectional attention on unmasked contexts. This innovative approach empowers it with heightened proficiency in both comprehension and generative tasks.

\vpara{Pre-Training Objectives.}  To amplify its linguistic understanding and generative abilities, GLM incorporates a dual pre-training strategy: 1) \textit{Autoregressive Blank Infilling} involves predicting missing tokens within spans of corrupted text, wherein segments are arbitrarily supplanted with a [MASK] token. 2) \textit{Multi-Task Pretraining} is utilized to endow GLM text generation ability, which aims to generate longer text by sampling random-length span from document-level or sentence-level text.

\vpara{Model Sizes.} GLM offers a diverse of models with various model parameters, including GLM-Large, GLM-6B, GLM-10B, GLM2-6B, ChatGLM-6B, and ChatGLM2-6B. Comprehensive specifics concerning the hyperparameters for each model variant can be found in Table~\ref{tab:glm_size}. GLM-Large model is specifically tailored for Chinese language processing tasks equipped with 335M model parameters, while GLM-10B, GLM-6B, and GLM2-6B are equipped with 10 billion, 6 billion, and 6 billion parameters, respectively, enabling them to handle a wide range of NLP tasks with varying complexities. Augmenting the series are bilingual conversational models: ChatGLM-6B and ChatGLM2-6B, both tailored for Chinese-English bilingual dialogue tasks. The ChatGLM-6B model, having 6.2 billion parameters, undergoes fine-tuning using Chinese Q\&A and dialogue datasets. In contrast, ChatGLM2-6B emerges as an evolved iteration of ChatGLM-6B, marking enhancements in performance, extended context handling, optimized inference, and broader applicability.

\begin{small}
    \begin{table}[h]
    \centering
    \renewcommand{\arraystretch}{1.15}
    \setlength{\tabcolsep}{1.5mm}{
    \begin{tabular}{c|cccc}  
    \toprule
     Model     & Dimension   & Heads & Layers & Parameters  \\
    \midrule
    GLM-Large-Chinese & 1024 & 24 & 16 & 335M \\
    GLM-10B-Chinese & 4096 & 64 & 48 & 10B  \\
    GLM-6B & 4096 & 32 & 28 & 6.2B   \\
    GLM2-6B & 4096 & 32 & 28 & 6.2B  \\
    \hline
    ChatGLM-6B & 4096 & 32 & 28 & 6.2B  \\
    ChatGLM2-6B & 4096 & 32 & 28 & 6.2B \\
    \bottomrule
    \end{tabular}} 
    \vspace{2mm}
    \caption{Hyperparameters of the backbone models.}
    \vspace{-0.3cm}
    \label{tab:glm_size}
    \end{table}
\end{small}

\subsection{Arithmetic Dataset}\label{appendix: dataset}
The training dataset for pre-training arithmetic model is created with a Python script.  The dataset includes a variety of arithmetic expressions, encompassing different types of arithmetic operations such as addition, subtraction, multiplication, division, and exponentiation. Each expression in the dataset is composed of various types of numbers, including integers, decimals, fractions, percents, and negative numbers. 
The training dataset consists of approximately 50 million arithmetic sequences. To investigate the impact of dataset scale on the arithmetic performance, we also create multiple datasets of varying sizes, including 1 million, 5 million, 10 million, and 25 million. This diverse representation of numbers ensures that the model can handle a wide range of numerical formats encountered in real-world arithmetic problems. 

To facilitate the learning of underlying calculation rules, the arithmetic expressions are designed to be more complex than simple two-number calculations. Instead, each expression in the dataset involves multiple steps of calculations, ranging from 2 to 10 steps. By creating multi-step expressions, the model is exposed to more intricate mathematical reasoning and is better equipped to handle complex arithmetic problem-solving. The details of expressions is presented as follows. Table~\ref{tab:examples from arithmetic datset} demonstrates examples from the arithmetic dataset.

\begin{itemize}
    \item Operations involving integers up to 10,000 that combine addition, subtraction, multiplication, and division.

    \item Exponentiation tasks using an integer base up to 10,000 and an integer exponent capped at 100.

    \item Bracketed expressions that include integers up to 10,000, combined with operations such as addition, subtraction, multiplication, and division.

    \item Lengthy arithmetic expressions that incorporate brackets and blend various numerical types, including integers, decimals, percentages, and negative numbers. These sequences utilize operations such as addition, subtraction, multiplication, and division.

    \item Arithmetic expressions involving fractions combined with various operations, including addition, subtraction, multiplication, and division.

\end{itemize}

\begin{small}
\begin{table*}[hbpt]
    \centering
    \renewcommand{\arraystretch}{1.35}
    \setlength{\tabcolsep}{0.1mm}{
    \small
    \begin{tabular}{p{4.3cm}p{9.3cm}}  
    \toprule
     Types &  Arithmetic Expression    \\
    \midrule 
    \multirow{6}{*}{Integre mixing operation} & 1+8/1*10+2=1+8*10+2=1+80+2=81+2=83 \\
    ~ & 53-2+23+51*56=53-2+23+2856=51+23+2856=74+2856=2930 \\
    ~ & 214-792*509*260*556=214-403128*260*556=214-104813280*556=214-58276183680=-58276183466 \\
    ~ & 1912*6800*6022-7250-1624=13001600*6022-7250-1624=78295635200-7250-1624=78295627950-1624=78295626326 \\
     \hline
     \multirow{3}{*}{Exponentiation} & 5170$\textasciicircum$0=1, 1$\textasciicircum$8756=1  \\
     ~ & 3$\textasciicircum$9=19683, 93$\textasciicircum$18=270827695297250208363869180422467849 \\
     ~ & 100$\textasciicircum$13=100000000000000000000000000 \\
     \hline
     \multirow{3}{*}{Expression of fractions} & ((49/24)*-(8/70))/-(34/80)=(+(49/24)*(8/70))/(34/80)=(392/1680)/(34/80)=\\
     ~ & (7/30)/(34/80)=(7/30)*(80/34)=(560/1020)=28/51 \\
     ~ &  (9947/9276)+(4411/9276)=14358/9276=2393/1546 \\
      \hline
     \multirow{2}{*}{Expression with brackets} & -7805+(4383/7377)=-7805+0.5941439609597398=-7804.40585603904 \\
     ~ & 8371*(-1945+8878)=8371*(-1945+8878)=8371*6933=58036143 \\
     \hline
     \multirow{5}{*}{Lengthy arithmetic expressions} & (-2090-5457.35697)*73.0=-7547.35697*73.0=-550957.05881 \\
     ~ & -4457+(-7823/5483\%)*-3338=-4457+(-7823/54.83)*-3338=-4457+(-142.6773664052526)*-3338=-4457+-142.6773664052526*-3338=-4457+142.6773664052526*3338=-4457+476257.0490607332=471800.0490607332 \\
    \bottomrule
    \end{tabular}} 
    \caption{Examples from the arithmetic dataset where ``+'', ``-'', ``*'', ``/'', ``$\textasciicircum$'' denotes addition, subtraction, multiplication, division, and exponentiation respectively. }
    \vspace{-0.6cm}
    \label{tab:examples from arithmetic datset}
\end{table*}
\end{small}

\subsection{Results on MATH 401}\label{appendix: results_in_math401}
Table~\ref{tab:arithmetic_results_all} shows a comprehensive evaluation of the arithmetic performance of \model on the MATH 401 dataset~\cite{yuan2023well}. This dataset offers a new set of arithmetic problems, allowing for a deeper exploration into \model's proficiency in addressing a wide variety of arithmetic tasks. By evaluating \model's performance on this dataset, we observe that \model consistently outperforms all other large language models with a substantial number of model parameters.

\begin{small}
    \begin{table}[thbp]
    \centering
    \renewcommand{\arraystretch}{1.15}
    \setlength{\tabcolsep}{2mm}{
    \begin{tabular}{c|cc}  
    \toprule
     Model     & ACC  &  \\
    \midrule
    GPT-4               & 83.54\% \\
    GPT-3.5-turbo       & 75.06\% \\
    text-davinci-003    & 56.61\%\\
    text-davinci-002    & 42.89\%  \\
    code-davinci-002    & 21.70\% & \\
    \hline      
    \hline
    Galactica-120b & 45.14\% & \\
    Galactica-30b  & 45.14\% & \\
    Galactica-6.7b & 34.41\% & \\
    \hline 
    \hline
    LLaMA-65b   & 28.43\% \\
    LLaMA-30b   & 30.17\% \\
    LLaMA-13b   & 27.68\% \\
    LLaMA-7b    & 21.96\% \\
    \hline
    \hline
    OPT-175B    & 21.70\% \\
    OPT-66B     & 20.70\% \\
    OPT-30B     & 15.96\% \\
    OPT-13B     & 15.21\% \\
    OPT-6.7B    & 14.46\% \\
    \hline
    \hline
    BLOOM-176B & 22.44\% \\
    BLOOM-7.1B & 7.23\% \\
    BLOOM-3B & 4.24\% \\
    BLOOM-1.7B & 5.24\% \\
    \hline
    \hline
    GLM-130B & 25.94\% \\
    GLM-10B  & 14.96\% \\
    \hline
    \hline
    \model-0.5B & 85.48\% \\
    \model-2B & 89.44\% \\
    \bottomrule
    \end{tabular}} 
    \vspace{2mm}
    \caption{Overall performance comparison on various LLMs in term of Accuracy.}
    \label{tab:arithmetic_results_all}
    \end{table}
\end{small}

\subsection{Analysis on Arithmetic Errors}\label{appendix: cases on char_error}

Table~\ref{tab:error_analysis} provides some examples to analyze the failures of \model on performing arithmetic tasks. Through careful examination of these examples, we can observe several patterns and trends in the \model's errors. Firstly, \model appears to grapple with intricate arithmetic expressions, particularly those combining several operations and large numbers. For instance, the expression 14031528/742: the division of an 8-digit number by a 4-digit one proves problematic for \model, leading to miscalculations in the outcome. Secondly, \model tends to encounter difficulties when dealing with long sequences of numbers and operations. As the expression length increases, the model's ability to accurately perform arithmetic calculations diminishes, leading to inaccurate results. For example, expression involving multiplication among two large numbers like 3626 * 8919 and calculation with a decimal and large integer number like 1.610311 * 7691. These errors generated by \model usually have only one calculation result error, indicating that the \model's mistakes mainly occur at specific calculation steps rather than affecting the entire expression.

\begin{small}
\begin{table*}[hbpt]
    \centering
    \renewcommand{\arraystretch}{1.35}
    \setlength{\tabcolsep}{0.2mm}{
    \small
    \scalebox{0.9}{
    \begin{tabular}{p{3.5cm}p{6cm}p{6cm}}  
    \toprule
     Input &  OutPut & Ground Truth    \\
    \midrule 
    3468*4046/7424=  & 14031528/7424=18\textcolor{red}{89.901400862069}  & 14031528/7424=18\textcolor{red}{90.0226293103449} \\
    (3626*8919)/8861= & 323\textcolor{red}{3}0294/8861=3648.605574991536 & 323\textcolor{red}{4}0294/8861=3649.7341157882856 \\
    7715/4791*7691-1968*9155= & 1.610311*7691-1968*9155=12384.\textcolor{red}{801801}-1968*9155=12384.801801-18017040=-18004655.198199 & 1.610311*7691-1968*9155=12384.\textcolor{red}{9018993}-1968*9155=12384.9018993-18017040=-18004655.098100606 \\
    (4059+7011.8718)-4038.22*847.15907= & (4059+7011.8718)-4038.22*847.15907=11070.8718-4038.22*847.15907=11070.8718-342\textcolor{red}{0}014.6996554=-3408943.8278554 & (4059+7011.8718)-4038.22*847.15907=11070.8718-4038.22*847.15907=11070.8718-342\textcolor{red}{1}014.6996554=-3409943.8278554003 \\
    7499-5747.91007/-5438*-439= & 7499-5747.91007/5438*439=7499-1.05\textcolor{red}{70081040823832}*439=7499-464.0265576921662=7034.973442307834 & 7499-5747.91007/5438*439=7499-1.05\textcolor{red}{6989715}*439=7499-464.0184848713=7034.981515128724 \\
    3868*6735*5755+3741-7533= & 26050980*5755+3741-7533=1\textcolor{red}{5}9923389900+3741-7533=159923393641-7533=159923386108 & 26050980*5755+3741-7533=1\textcolor{red}{4}9923389900+3741-7533=149923393641-7533=149923386108 \\
    \bottomrule
    \end{tabular}}} 
    \caption{Some failed examples generated by \model. }
    \label{tab:error_analysis}
\end{table*}
\end{small}

\subsection{K6 Dataset}\label{appendix: k12 dataset}

We collect math word problems from Chinese elementary schools in collaboration with the renowned educational institution, TAL AI Lab. The dataset consists of math problems for each grade level, with each grade containing approximately 100 problems. The wide-ranging nature of these math word problems empowers us to gauge the model's efficacy across an array of difficulty gradients and academic grades.
To illustrate the diversity and complexity of the K6 dataset, we present some exemplary math word problems in Table~\ref{tab:examples_k12}. These examples show the range of mathematical concepts covered and the varying levels of difficulty present in the dataset.

\begin{small}
\begin{table*}[hbpt]
    \centering
    \renewcommand{\arraystretch}{1.35}
    \setlength{\tabcolsep}{0.1mm}{
    \small
    \begin{tabular}{p{2cm}p{12cm}}  
    \toprule
     Grade &  Example    \\
    \midrule 
    K1 & \begin{CJK}{UTF8}{gbsn}李老师买了20颗糖果,送给小丽5颗,送给小刚8颗,还剩多少颗糖果? \end{CJK} \\
    K2 & \begin{CJK}{UTF8}{gbsn}一个乘数是4,另一个乘数是7,积是多少?\end{CJK} \\
    K3 & \begin{CJK}{UTF8}{gbsn}乐乐家养了36只小鸡,其中1/4是公鸡,母鸡是公鸡的3倍,公鸡和母鸡各有多少只?\end{CJK}  \\
    K4 & \begin{CJK}{UTF8}{gbsn}公益小组的同学为敬老院的老人们制作香囊(náng ),12个组共制作了864个,每组都有9人,平均每人制作了几个?\end{CJK}  \\
    K5 & \begin{CJK}{UTF8}{gbsn}东、西两城相距180千米,甲、乙两车分别从东、西两城同时出发,相向而行,1.2小时后两车可相遇.实际甲车出发0.4小时后因故障停车,乙车又走了2小时才和甲车相遇,求乙车每小时行多少千米? \end{CJK} \\
    K6 & \begin{CJK}{UTF8}{gbsn}甜甜读一本小说,第一天读了这本书的3/8,正好是180页,第二天又读了这本书的1/6,第2天读了多少页?\end{CJK} \\
    \bottomrule
    \end{tabular}} 
    \caption{Examples from the K6 dataset to demonstrate the diversity and complexity of this dataset. }
    \label{tab:examples_k12}
\end{table*}
\end{small}

\subsection{Baseline Models}\label{appendix:baseline_models}

Here, we leverage a variety of popular LLMs that can address Chinese problems to compare the mathematical reasoning ability among these LLMs and our \model. The details of each baseline LLM as follows.

\begin{itemize}
    \item GPT-4~\cite{2303.08774} is the most advanced generative language model that developed by OpenAI, which successfully achieves so many SOTA performances on a variety of downstream tasks. 
    \item ChatGPT~\cite{chatgpt} is the predecessor of GPT4 and is constructed upon the success of InstructGPT~\cite{ouyang2022training}, which is fine-tuned using instruction data with reinforcement learning from human feedback (RLHF), making it a powerful tool for natural language understanding and conversation.
    \item MOSS~\cite{moss} is an open-source LLM that consists of 16 billion model parameters. It  utilizes 100 billion Chinese tokens and 20 billion English tokens to learn language patterns and semantic representations.
    \item Ziya-LLaMA-13B~\cite{fengshenbang} is a language model constructed on LLaMA-13B, 
    which extends LLaMA-13B's character set to contain 7,000 Chinese characters and undergoes continual pre-training on a vast dataset of 110 billion Chinese tokens.
    \item Chinese-Alpaca-13B~\cite{cui2023efficient} is a Chinese language model with 13 billion parameters that is built upon LLaMA-13B. During the supervised instruction tuning, the Low Rank Adaptation (LoRA)~\cite{hu2021lora} technique is utilized to fine-tune LLaMA-13B for Chinese language tasks.
    \item Baichuan-7B~\cite{baichuan} shares similarities with LLaMA but is pre-trained from scratch on a massive dataset containing 1.2 trillion Chinese and English tokens. 
    \item ChatGLM-6B~\cite{chatglm} and its successor ChatGLM2-6B~\cite{chatglm2} are language models that share a unified transformer architecture named GLM~\cite{du2021glm,zeng2022glm}. These models are pre-trained on a diverse dataset containing English and Chinese data, combined with the supervised instruction tuning, makes them powerful tools for understanding and generating text in both English and Chinese contexts.
\end{itemize}

\subsection{Training Steps Analysis.} 
\begin{figure*}[htbp]
    \centering
    \includegraphics[width=\textwidth]{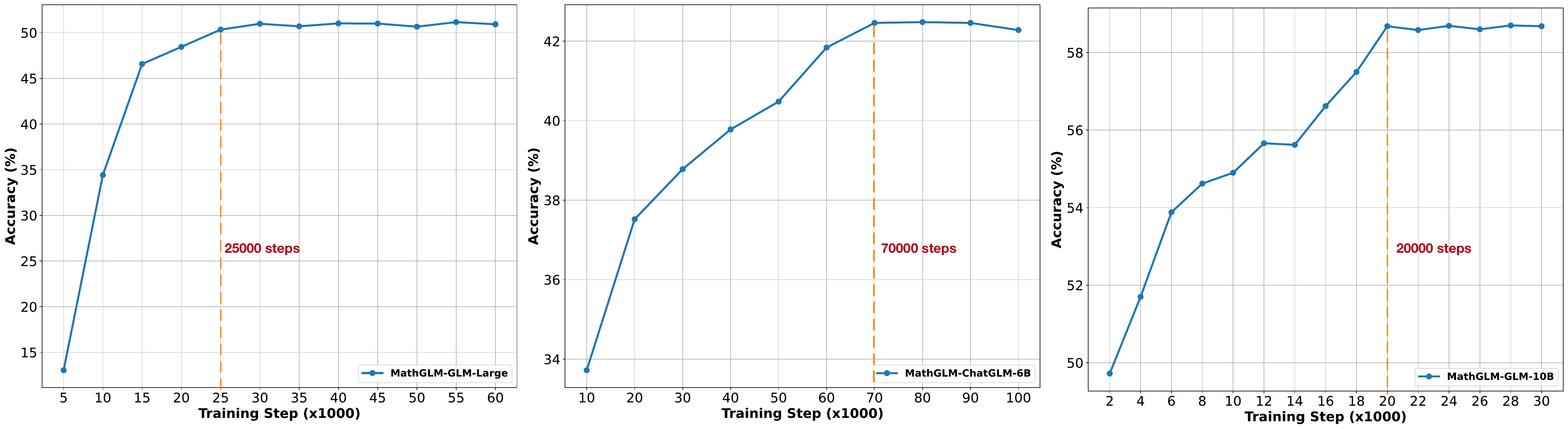}
    \caption{The impact of training steps on \model with different backbone models. Fine-tuning on model undergoing instruction tuning requires a longer training steps.
    }
    \label{fig:training_steps}
\end{figure*}

We explore the impact of training steps on the \model's performance by analyzing its performance against varied training steps, as depicted in Figure~\ref{fig:training_steps}. The results reveal that there is a consistent uptrend in performance as the number of training steps increases. With more training steps, \model becomes increasingly adept at comprehending and resolving math word problems, which translates to a surge in accuracy. However, it is clearly observed that the performance gains of \model start to plateau after a certain point, indicating potential diminishing returns with extended training. These findings highlight the significance of finding an optimal balance between training time and performance gains for \model in solving math word prblems. Additionally, we observe that model undergoing instruction tuning requires a longer training duration to achieve consistent accuracy on math word problems.

\subsection{Case study}
\begin{figure}[htbp]
    \centering
    \includegraphics[width=\textwidth]{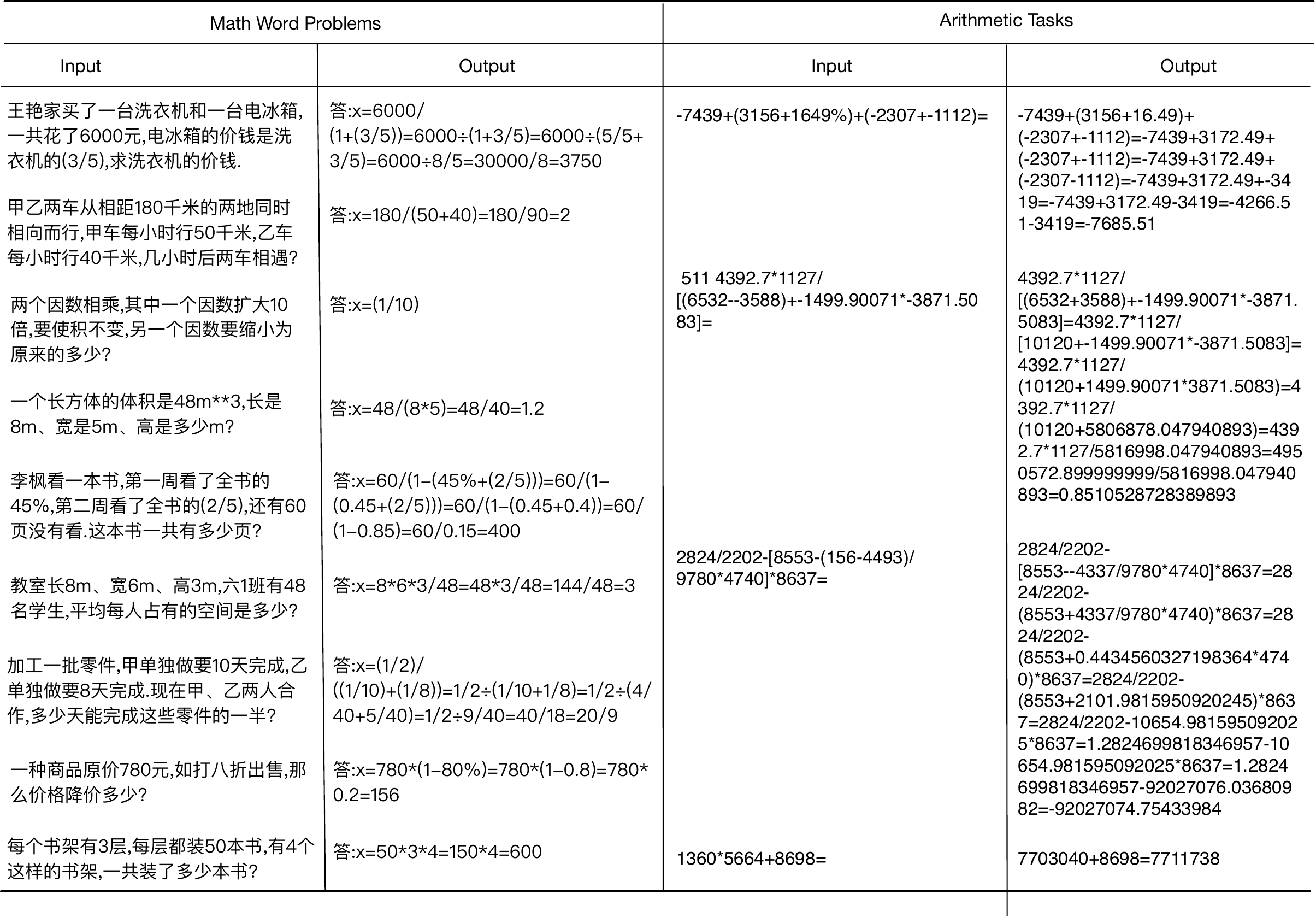}
    \caption{Some cases generated by \model on arithmetic tasks and math word problems.
    }
    \label{fig:cases}
\end{figure}

Here, we present specific cases to demonstrate the solving process of the \model on both arithmetic tasks and math word problems. As shown in Figure~\ref{fig:cases}, these examples illustrate how the \model leverages a step-by-step strategy to solve both arithmetic tasks and math word problems accurately.

For arithmetic tasks, the \model breaks down complex calculations into multiple steps rather than a direct answer. The step-by-step strategy ensures that each intermediate result is accurately computed, leading to the final correct answer for each arithmetic expression.

In math word problems, the \model utilizes LLMs's language understanding capabilities to understand the problem statement accurately. The \model then applies the step-by-step strategy to seperately calculate the generated arithmetic expressions for each math problem, enabling it to obtain the correct answer for math word problems.

\end{document}